%% file: main.tex
\documentclass[letterpaper]{article} %
\usepackage{aaai2026}  %
\usepackage{times}  %
\usepackage{helvet}  %
\usepackage{courier}  %
\usepackage[hyphens]{url}  %
\usepackage{graphicx} %
\usepackage{natbib}  %
\usepackage{caption} %
\usepackage{algorithm}
\usepackage{algorithmic}
\usepackage{amsfonts}
\usepackage{amsmath}
\usepackage{amsthm}
\usepackage{subcaption}

\usepackage{wasysym}

\usepackage{booktabs} 
\usepackage{multicol}
\usepackage{multirow}

\usepackage{times}
\usepackage{helvet}
\usepackage{courier}
\usepackage{xcolor}
\usepackage{comment}
\usepackage{newfloat}
\usepackage{listings}
\DeclareCaptionStyle{ruled}{labelfont=normalfont,labelsep=colon,strut=off} %
\floatstyle{ruled}
\newfloat{listing}{tb}{lst}{}
\floatname{listing}{Listing}
\nocopyright
\usepackage{hyperref} 

\title{iFairy: the First 2-bit Complex LLM with All Parameters in $\{\pm1, \pm i\}$}

\author {
    Feiyu Wang,
    Guoan Wang,
    Yihao Zhang,
    Shengfan Wang,
    Weitao Li,
    Bokai Huang,\\
    Shimao Chen,
    Zihan Jiang,
    Rui Xu,
    Tong Yang\footnote{Corresponding author: yangtong@pku.edu.cn}
}
\affiliations {
    Peking University\\
    \textbf{Model-700M: }\url{https://huggingface.co/PKU-DS-LAB/Fairy-plus-minus-i-700M}\\
    \textbf{Model-1.3B: }\url{https://huggingface.co/PKU-DS-LAB/Fairy-plus-minus-i-1.3B}\\
    \textbf{Code: }\url{https://github.com/PKULab1806/Fairy-plus-minus-i}
}

\usepackage{bibentry}

\begin{document}
\input{body/macro}

\maketitle

\input{body/0asb_yt}

\input{body/1intro_wfy_version1}
\input{body/2relate}
\input{body/3model}
\input{body/4eval}
\input{body/5conclusion}

\clearpage
\bibliography{mybib.bib}

\input{body/appendix}

\end{document}

%% file: body/macro.tex
\newcommand{\mname}{\textit{i}Fairy}
\newcommand{\clinear}{ComplexLinear}
\newcommand{\qscheme}{PhaseQuant}

%% file: body/0asb_yt.tex
\begin{abstract}

Quantization-Aware Training (QAT) integrates quantization into the training loop, enabling LLMs to learn robust low-bit representations, and is widely recognized as one of the most promising research directions.
All current QAT research focuses on minimizing quantization error on full-precision models, where the full-precision accuracy acts as an upper bound (accuracy ceiling).
No existing method has even attempted to surpass this ceiling.
To break this ceiling, we propose a new paradigm: raising the ceiling (full-precision model), and then still quantizing it efficiently into 2 bits.
We propose \mname{}, the first 2-bit quantization framework for complex-valued LLMs. 
Specifically, our method leverages the representational advantages of the complex domain to boost full-precision accuracy. We map weights to the fourth roots of unity $\{\pm1, \pm i\}$, forming a perfectly symmetric and information-theoretically optimal 2-bit representation. 
Importantly, each quantized weight has either a zero real or imaginary part, enabling multiplication-free inference using only additions and element swaps.
Experimental results show that \mname{} outperforms the ceiling of existing 2-bit quantization approaches in terms of both PPL and downstream tasks, while maintaining strict storage and compute efficiency.
This work opens a new direction for building highly accurate and practical LLMs under extremely low-bit constraints.
\end{abstract}

%% file: body/1intro_wfy_version1.tex
\section{Introduction}

The advent of Large Language Models (LLMs) has transformed artificial intelligence, achieving remarkable performance across a wide range of natural language tasks~\cite{achiam2023gpt, touvron2023llama2, dubey2024llama3}. However, the deployment of these models in real-world applications faces two critical bottlenecks: \textbf{spatial} and \textbf{temporal}. The spatial bottleneck arises from the massive number of parameters, which is often in the billions or trillions, leading to prohibitive storage requirements and large memory footprints. The temporal bottleneck, on the other hand, stems from the heavy reliance on large-scale matrix multiplications during inference, which not only slows down computation but also significantly increases power consumption. Overcoming these two bottlenecks by maintaining model accuracy under extreme compression and reducing or eliminating costly multiplications would greatly enhance the efficiency of LLMs, enabling transformative applications in domains such as physics, chemistry, biology, astronomy, and geoscience.

Model compression has thus become a critical research area, with \textit{quantization} emerging as one of the most promising techniques to alleviate these bottlenecks~\cite{miao2023inferenceSurvey1, wan2023inferenceSurvey2}. Quantization methods are broadly categorized into Post-Training Quantization (PTQ) and Quantization-Aware Training (QAT). While PTQ~\cite{frantar2022gptq, lin2023awq} offers simplicity, its performance often degrades sharply in extremely low-bit scenarios due to the model’s lack of adaptation to quantized representations. In contrast, QAT integrates quantization into the training loop, allowing models to learn robust low-bit representations and maintain performance under aggressive compression. This advantage has motivated recent research into QAT-based strategies tailored for LLMs.

The pursuit of extremely low-bit quantization, particularly 2-bit quantization, has become a focal point in efforts to compress Large LLMs for efficient deployment. Existing approaches, such as BitNet~\cite{wang2023bitnet} and its successors~\cite{ma2024bitnetb1.58}, have demonstrated that it is possible to retain reasonable accuracy using ternary quantization schemes with just 1.58 bits per weight. However, the accuracy of any quantized model is fundamentally limited by the following equation:
\[
\textbf{Accuracy}_\text{quant}=\textbf{Accuracy}_\text{full-precision}-\textbf{Error}_\text{quant}
\]
All current quantization research focuses on minimizing quantization error on full-precision models (e.g., LLaMA), but the quantization error can never be zero. Therefore, full-precision accuracy becomes the \textbf{ceiling} for quantized accuracy. To date, no existing method has even attempted to surpass this ceiling.

In this paper, we propose a fundamentally different perspective. Instead of solely focusing on reducing quantization error, we make the first attempt to raise the ceiling (the accuracy of the full-precision model), while still ensuring that the resulting model can be efficiently quantized to a 2-bit format. Our key insight is that if the full-precision model becomes more expressive and accurate, the final 2-bit quantized model can achieve higher accuracy as well.
Building on this insight, we propose, for the first time, incorporating complex-valued neural architectures into LLMs. The complex number provides a richer representational space with additional phase information, thereby enhancing the expressiveness of linear transformations without increasing the parameter count. By systematically extending the Transformer architecture into the complex domain, we construct a full-precision complex-valued LLM with superior modeling capacity.

Building upon this complex-valued foundation, we further design a novel 2-bit quantization scheme tailored for complex weights. Specifically, we quantize each complex parameter to one of the \textbf{fourth roots of unity} $\{\pm 1, \pm i\}$ in the complex plane. This approach, unlike real-valued quantization, exploits the full 2-bit representational capacity \textit{without sacrificing symmetry or sparsity}, thereby eliminating the trade-offs that limit real-valued schemes. The resulting model, which we name \mname{} (also named Fairy $\pm i$, Fairy-imaginary), is perfectly storage-efficient and phase-aware by design.
We propose a quantization function \qscheme{} that learns to project full-precision complex weights onto the target set $\{\pm 1, \pm i\}$ while preserving both magnitude and phase information. We implement this within our complex Transformer framework and evaluate its performance under the same storage and compute constraints as BitNet b1.58. Experiments show that \mname{} significantly improves perplexity and downstream task accuracy, outperforming existing 2-bit baselines and approaching the performance of full-precision FP16 models.

Our contributions can be summarized as follows:
\begin{itemize}
    \item We propose a new perspective on low-bit quantization: improving the accuracy of quantized models by raising the ceiling (the full precision model).
    \item We design a complex-valued LLM architecture that leverages the representational benefits of the complex domain without increasing parameter storage.
    \item We design a 2-bit quantization scheme that maps complex weights to the 4th roots of unity $\{\pm 1, \pm i\}$, fully utilizing bit capacity while preserving key properties like symmetry and sparsity.
    \item Experimental results show that our quantized model outperforms the ceiling of existing 2-bit quantization approaches in terms of both PPL and downstream understanding tasks.
\end{itemize}

%% file: body/2relate.tex
\section{Related Work}
\subsection{Quantization Techniques} 
The effort to mitigate the computational burden of large language models has led to significant research in model quantization. Quantization aims to reduce memory footprint and computation cost by representing weights, activations, and occasionally gradients using low-precision numerical formats. Existing quantization methods are typically categorized into post-training quantization (PTQ) and quantization-aware training (QAT).
Post-training quantization (PTQ) applies quantization to a pretrained full-precision model without additional training. Notable methods include GPTQ~\cite{frantar2022gptq}, a one-shot quantization algorithm that leverages approximate second-order information; AWQ~\cite{lin2023awq}, introducing channel-wise weight quantization along with activation weighting to improve output calibration; and SmoothQuant~\cite{xiao2023smoothquant}, which jointly scales weights and activations to enable robust 8-bit quantization. These PTQ techniques have shown remarkable performance with minimal degradation.
Quantization-aware training (QAT), in contrast, integrates quantization directly into the training process. During both forward and backward passes, quantized values are used to allow the model to adapt to quantization-induced constraints. QAT generally yields better accuracy than PTQ, especially for sub-4-bit precision and end-to-end quantized models. While QAT introduces additional training overhead, it enables more accurate models. Typical QAT works include~\cite{liu2023llmqat, chen2024efficientqat, bondarenko2024lowrankqat}.
In this work, we propose a novel QAT-based quantization framework specifically designed for extremely low-bit complex-valued language models.

\subsection{Extremely Low-Bit LLMs}
To reduce the storage and computational costs of large-scale models, many studies have extended binary neural network techniques to large language models. Early works on binary neural networks such as BinaryConnect~\cite{courbariaux2015binaryconnect}, BinaryNet~\cite{courbariaux2016binarized}, and XNOR-Net~\cite{rastegari2016xnor} proposed binarizing weights to \(\{-1, +1\}\), using the Straight-Through Estimator (STE)~\cite{bengio2013ste} to enable training.
BitNet~\cite{wang2023bitnet} scaled this idea to LLMs by introducing \textit{BitLinear} layers with binary weights, enabling addition-only inference while preserving competitive accuracy. BitNet b1.58~\cite{ma2024bitnetb1.58} further extended the weight set to ternary \(\{-1, 0, +1\}\), improving expressiveness under the same 2-bit budget. Subsequent variants~\cite{wang2025bitnetv2, wang2024bitneta4.8, ma2025bitnet2B4T} and related efforts~\cite{team2025minicpm4} further advanced the practical deployment of extremely low-bit LLMs.
More recently, ParetoQ~\cite{liu2025paretoq} explores scaling laws in extremely low-bit LLM Quantization
These works inspire our approach: by extending Transformer architectures into the complex domain and adopting a symmetric, phase-aware 2-bit quantization scheme, we aim to overcome the representational inefficiencies of real-valued BitNet design while preserving their computational advantages.

\subsection{Complex-Valued Neural Networks}
The use of complex numbers in neural networks is not a new concept. Complex-Valued Neural Networks (CVNNs) have been explored for several decades, particularly in domains where data possesses inherent phase and magnitude properties, such as in signal processing and imaging~\cite{hirose2006complex, lee2022complex-survey, bassey2021survey-survey2, yang2020complextransformer, eilers2023complextransformer2}. By representing weights and activations as complex numbers, CVNNs can potentially capture more intricate patterns and feature dependencies compared to their real-valued counterparts. However, the application of CVNNs to natural language processing, and specifically to LLMs, has been limited. Our work bridges this gap by demonstrating that the complex domain offers a compelling solution to a fundamental efficiency problem in 1-bit real-valued quantization.

In contrast to existing approaches, our work integrates the advantages of complex-valued representations with efficient, extremely low-bit quantization strategies, bridging a crucial gap and opening new avenues for more efficient deployment of powerful language models.

%% file: body/3model.tex
\section{The \mname{} model}
\begin{figure*}[ht]
    \centering
    \begin{subfigure}[b]{0.49\textwidth}
        \includegraphics[width=\linewidth]{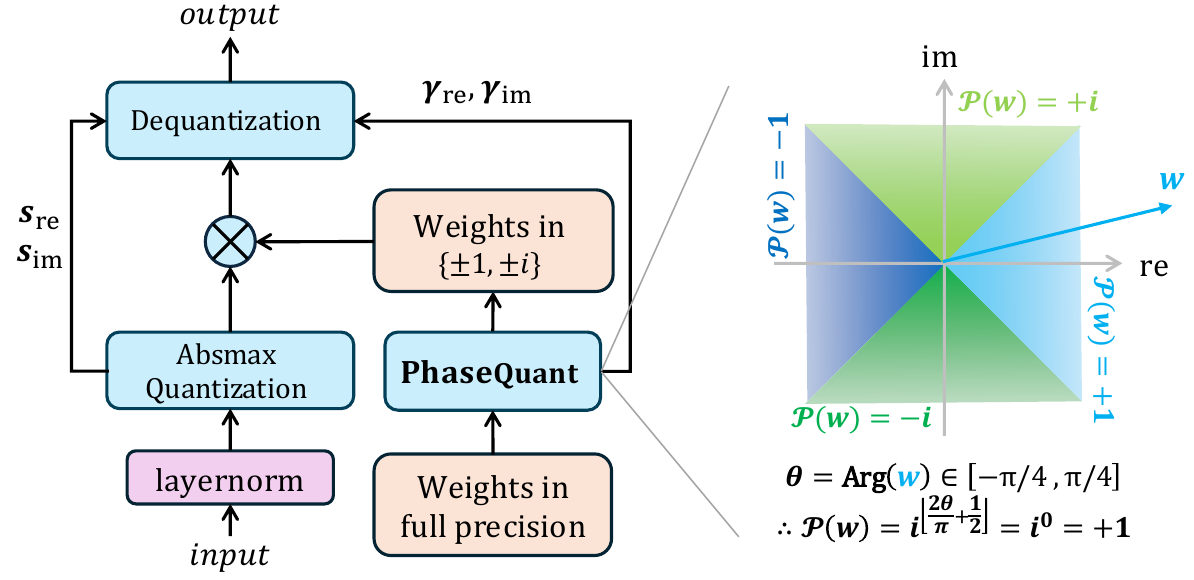}
        \caption{\clinear{} with \qscheme{}.}
        \label{fig:complexquant}
    \end{subfigure}
    \hspace{0.04\textwidth}
    \begin{subfigure}[b]{0.41\textwidth}
        \includegraphics[width=\linewidth]{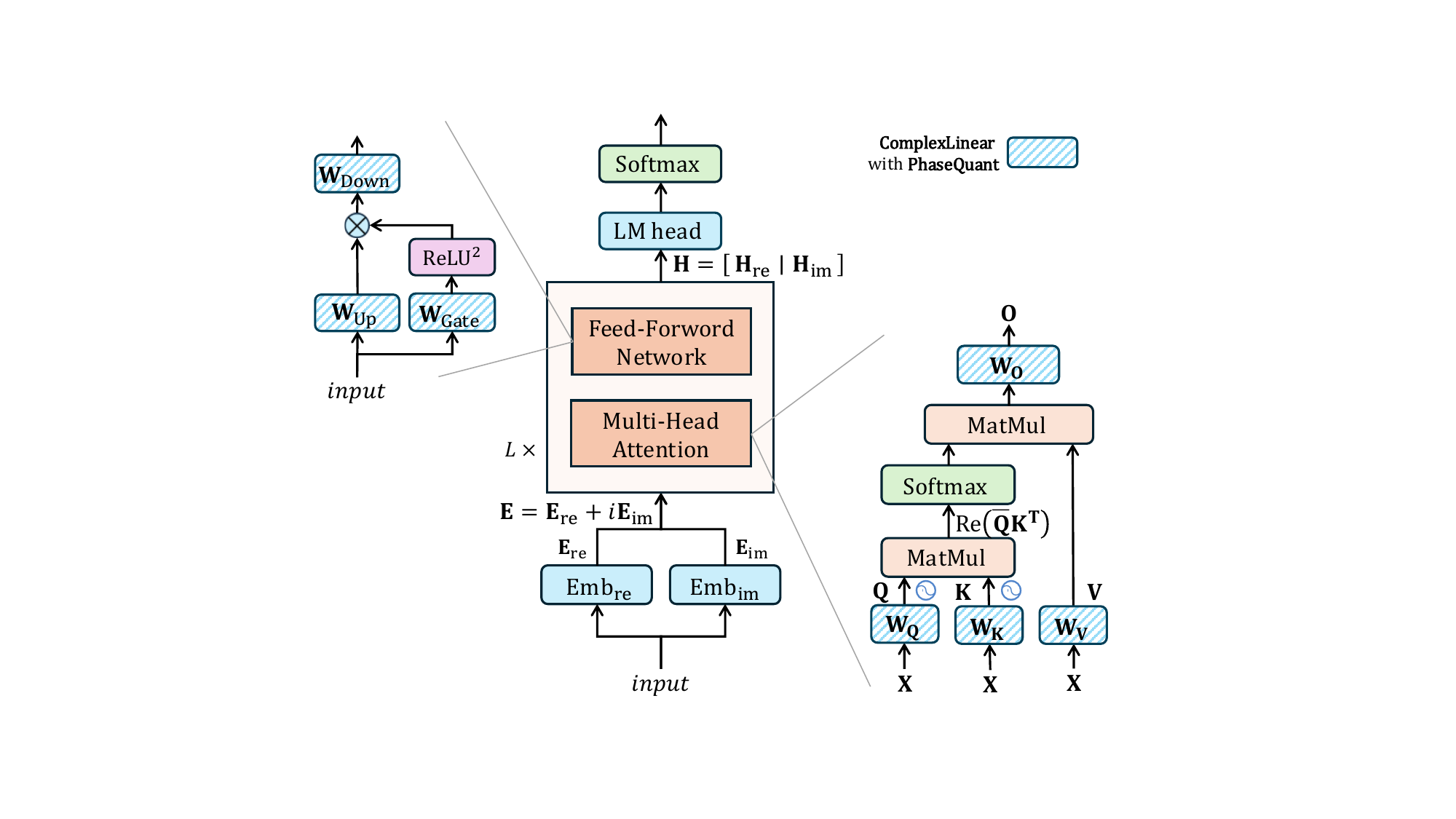}
        \caption{\mname{} backbone.}
        \label{fig:complexbitnet}
    \end{subfigure}
    \caption{Overview of \qscheme{} and \mname{}. The left panel illustrates the quantization process of \qscheme{}. In the right panel, \qscheme{} is applied to all major linear projections within \mname{}, including $\mathbf{W}_\mathbf{Q}$, $\mathbf{W}_\mathbf{K}$, $\mathbf{W}_\mathbf{V}$, and $\mathbf{W}_\mathbf{O}$ in the self-attention block, as well as $\mathbf{W}_\text{Up}$, $\mathbf{W}_\text{Gate}$, and $\mathbf{W}_\text{Down}$ in the feed-forward network.}

    \label{fig:overview}
\end{figure*}

In this section, we propose \mname{}, which extends the conventional Transformer architecture into the complex domain, and enables more expressive and efficient representations through the use of native 2-bit complex-valued weights. 
This section is structured as follows: Section~\ref{subsec:model_backbone} details the architectural adaptations required for complex-valued operations within the Transformer backbone. Section~\ref{subsec:weight_quant} and \ref{subsec:act_quant} then describe our quantization strategies for complex-valued weights and activations, respectively. Finally, Section~\ref{subsec:complexity} compares the computational complexity of \mname{}, BitNet1.58, and the full-precision Llama model.

\subsection{Model Architecture of \mname{}}
\label{subsec:model_architecture}

Our proposed model, \mname{}, is a highly efficient Transformer designed to operate with 2-bit complex-valued weights. It leverages the rich representational capacity of the complex domain while maintaining the computational benefits of extremely low-bit quantization. The overall architecture is illustrated in Figure~\ref{fig:overview}.
The foundation of \mname{} is a Complex-Valued Transformer Backbone. This backbone adapts a standard LLaMA-style architecture to the complex domain, re-engineering core components, such as the embedding layers, self-attention layers, language model head, and feed-forward networks, with \clinear{} module, to handle complex-valued parameters and activations. 
This design provides the expressive power needed to learn complex data patterns. 
We then propose our primary \qscheme{} scheme to map the complex-valued weights of the backbone into a discrete 2-bit complex space during computation. 
The \mname{} is constructed by systematically applying \qscheme{} to \clinear{} in the complex-valued Transformer model, as shown on the right of Figure~\ref{fig:overview}.

\subsection{Backbone of \mname{}}
\label{subsec:model_backbone}
The foundational principle of our methodology is the systematic extension of the Transformer architecture to operate on complex numbers. Specifically, all model parameters and intermediate representations are complex values. A weight matrix \( \mathbf{W} \in \mathbb{C}^{m \times n} \) and an input activation \( \mathbf{x} \in \mathbb{C}^{m} \) are thus represented by their real and imaginary parts:
\[
\mathbf{W} = \mathbf{W}_{\text{re}} + i\mathbf{W}_{\text{im}},
\]
\[
\mathbf{x} = \mathbf{x}_{\text{re}} + i\mathbf{x}_{\text{im}},
\]
where \( \mathbf{W}_{\text{re}}, \mathbf{W}_{\text{im}} \in \mathbb{R}^{m \times n} \), and \( \mathbf{x}_{\text{re}}, \mathbf{x}_{\text{im}} \in \mathbb{R}^{m} \). Incorporating the mathematically robust properties of positive definiteness and conjugate symmetry, the Hermitian inner product is naturally employed for linear projection in the backbone. It is defined as:
\[\mathbf{Y}=\overline{\mathbf{x}} \mathbf{W},\]
where $\overline{\mathbf{x}}$ denotes the complex conjugate of $\mathbf{x}$. 
We designate the layer that performs this operation as the \clinear{}, which serves as the complex-valued counterpart to the standard linear layer.
The following subsections detail the primary modifications made to the standard Transformer architecture to realize this complex-valued backbone. A more detailed justification for these architectural choices is provided in Appendix~\ref{appendix:justification}.

\subsubsection{Dual-channel Projection Embedding Layers.}
\begin{figure}[ht]
    \centering
    \begin{subfigure}[b]{0.195\textwidth}
        \includegraphics[width=\linewidth]{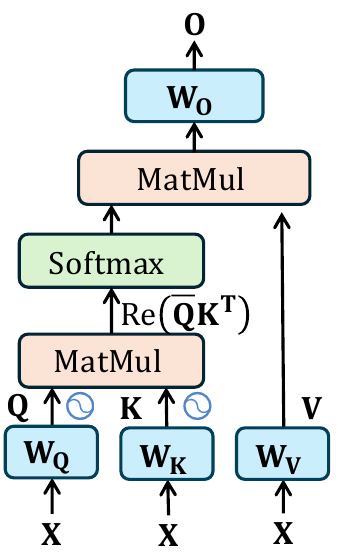}
        \caption{Attention mechanism.}
        \label{fig:attn}
    \end{subfigure}
    \hspace{0.02\textwidth}
    \begin{subfigure}[b]{0.245\textwidth}
        \includegraphics[width=\linewidth]{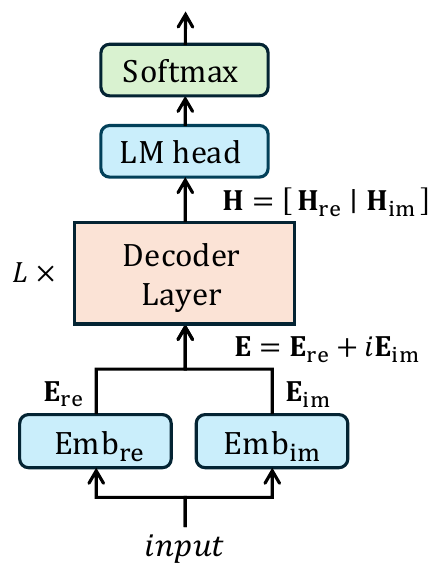}
        \caption{Embedding and LM head.}
        \label{fig:emb&lmhead}
    \end{subfigure}
    \caption{The complex-valued Transformer architecture.}
    \label{fig:all}
\end{figure}

To bridge the discrete token space and the continuous complex-valued representation space, we employ a dual-channel projecting strategy, as illustrated in the bottom part of Figure~\ref{fig:emb&lmhead}. This is implemented using two parallel embedding layers. For a given input token, one layer generates the real part of the embedding vector $\mathbf{E_\text{re}}$, while the other generates the imaginary part $\mathbf{E_\text{im}}$, forming the final complex embedding $\mathbf{E}=\mathbf{E_\text{re}}+i\mathbf{E_\text{im}}$. Although structurally separate, the two pathways are implicitly coupled through the subsequent complex-valued operations, which encourage the model to learn the unified complex representation during the end-to-end training.

\subsubsection{Efficient Complex-Valued Self-Attention.}
Extending the self-attention mechanism to the complex domain raises a fundamental question: What is a principled and computationally efficient way to define similarity between complex-valued queries and keys that also respects the geometric structure of complex vector spaces?

In our formulation, we adopt the real part of the Hermitian inner product as the attention score:
\[
S=\text{score}(\mathbf{Q}, \mathbf{K}) = \operatorname{Re}(\overline{\mathbf{Q}}\mathbf{K}^T) = \mathbf{Q}_\text{re}\mathbf{K}_\text{re}^T + \mathbf{Q}_\text{im}\mathbf{K}_\text{im}^T. 
\]
This choice ensures that all four components of $\mathbf{Q}$ and $\mathbf{K}$---real and imaginary parts of both---are involved, thus preserving informational completeness while maintaining compatibility with standard real-valued softmax operations. Crucially, this approach admits a well-established mathematical and geometric interpretation.
As discussed in~\cite{Scharnhorst:1999angles}, the real part of the Hermitian inner product corresponds to the so-called \emph{Euclidean angle} between complex vectors, defined by isometrically embedding the complex space $\mathbb{C}^n$ into $\mathbb{R}^{2n}$. See more in Appendix~\ref{appendix:justification}.

The final attention output $\mathbf{O}$ is then computed by applying the softmax function to the real-valued scores and multiplying the result with the complex-valued value vectors $\mathbf{V}$:
\[
\mathbf{O} = \text{softmax}\left(\frac{S}{\sqrt{d_k}}\right) \mathbf{V}.
\]
For practical efficiency, we recast the complex-valued computation as a larger real-valued matrix multiplication. We concatenate the real and imaginary parts of $\mathbf{Q}$, $\mathbf{K}$, and $\mathbf{V}$ along the last dimension to obtain:
\[
\tilde{\mathbf{Q}} = [\mathbf{Q}_\text{re} \mid \mathbf{Q}_\text{im}], \quad \text{w.l.o.g.} \quad \tilde{\mathbf{K}}, \tilde{\mathbf{V}}
\]
The score matrix can then be computed as:
$
S = \tilde{\mathbf{Q}} \tilde{\mathbf{K}}^\top.
$
This transformation enables the use of highly optimized real-valued FlashAttention kernels~\cite{dao2022flashattention, dao2023flashattention2}, while preserving the expressiveness of complex-valued representations. The resulting output is partitioned back into its real and imaginary components to construct the final complex-valued attention output $\mathbf{O}$. The entire procedure is summarized in Algorithm~\ref{alg:flash_attention}.
\begin{algorithm}[ht]
\caption{Efficient Complex-Valued Self-Attention}
\label{alg:flash_attention}
\begin{algorithmic}[1]
\STATE \textbf{Input:} Complex Query $\mathbf{Q}$, Key $\mathbf{K}$, Value $\mathbf{V}$
\STATE \textbf{Output:} Complex attention output $\mathbf{O}$
\STATE // Concatenate real and imaginary parts for \(\mathbf{Q}\), \(\mathbf{K}\), \(\mathbf{V}\) 
\FOR{each matrix \(\mathbf{M} \in \{\mathbf{Q}, \mathbf{K}, \mathbf{V}\}\)}
    \STATE \(\tilde{\mathbf{M}} \gets [\mathbf{M}_{\text{re}} \mid \mathbf{M}_{\text{im}}]\)
\ENDFOR
\STATE // Utilize standard Flash Attention kernel
\STATE $\tilde{\mathbf{O}} \gets \operatorname{FlashAttention}(\tilde{\mathbf{Q}}, \tilde{\mathbf{K}}, \tilde{\mathbf{V}})$
\STATE \(d \gets \text{dimension of } \tilde{\mathbf{O}} \text{ along the last axis}\)
\STATE \(\mathbf{O}_{\text{re}} \gets \tilde{\mathbf{O}}[ \dots, :d/2]\)
\STATE \(\mathbf{O}_{\text{im}} \gets \tilde{\mathbf{O}}[ \dots, d/2:]\)
\STATE \textbf{return} $\mathbf{O}$
\end{algorithmic}
\end{algorithm}

\subsubsection{Complex-Valued Feed-Forward Network.}
The Feed-Forward Network (FFN) in our architecture mirrors the structure of modern LLMs like LLaMA, but operates on complex numbers. 
A key modification resides in the non-linear activation function. For the activation function $f$, we use the squared ReLU~\cite{zhang2024relu2}, defined as $f(x)=\operatorname{ReLU}^2(x) = (\max(0, x))^2$. Let $\mathbf{Z} = \mathbf{Z}_\text{re} + i\mathbf{Z}_\text{im}$ is the gated activation, then the application of $f$ is:
$$ f(\mathbf{Z}) = \operatorname{ReLU}^2(\mathbf{Z}_\text{re}) + i \operatorname{ReLU}^2(\mathbf{Z}_\text{im}) $$
This design allows the network to maintain non-linearity, which is crucial for learning complex patterns, while confining the operation to the real domain where such functions are well-defined and computationally inexpensive.

\subsubsection{Complex Language Model Head.}
To project the final complex hidden states back to the vocabulary space for token prediction, we design our Language Model Head to be symmetric to the input embedding layer. This pattern ensures a principled and consistent mapping between the discrete token space and the continuous complex representation space at both ends of the network.
As illustrated at the top of Figure~\ref{fig:emb&lmhead}, the final hidden state $\mathbf{H} = \mathbf{H}_\text{re} + i\mathbf{H}_\text{im}$ is first transformed by concatenating its real and imaginary components into a single, larger real-valued matrix, $\tilde{\mathbf{H}} = [\mathbf{H}_{\text{re}} \mid \mathbf{H}_{\text{im}}]$. This matrix is then projected through a standard real-valued linear layer to compute the final logits over the vocabulary:
\[\text{logits}=\tilde{\mathbf{H}}\mathbf{W}^T_{out},\]
where $\mathbf{W}_{\text{out}}$ is the learned weight matrix of the output projection layer. 
This computational form creates a strong inductive bias, encouraging the output layer to learn a projection that measures the similarity between the final hidden state and each vocabulary token's latent representation using the same underlying metric that governs the model's internal reasoning. Consequently, the symmetric output projection provides a coherent and elegant solution for the final classification step. It ensures that the transformation from complex representations to token probabilities is not an arbitrary mapping, but one that is deeply integrated with the geometric principles established throughout the entire complex-valued architecture.

\subsubsection{Complex Rotary Position Embedding.}
In its original real-valued formulation, RoPE encodes absolute position by applying a rotation matrix to pairs of features in the query and key vectors. In the complex domain, this rotational logic can be implemented more directly, as a 2D rotation is equivalent to multiplication by a complex number of unit modulus, $e^{i\theta}$.
Given a token at position $m$ and a hidden dimension $j$, the rotary embedding is applied as follows:
\[ \mathbf{q}'_{m,j} = \mathbf{q}_{m,j} e^{i m \theta_j} ;\ \mathbf{k}'_{n,j} = \mathbf{k}_{n,j} e^{i n \theta_j} \]
where $\theta_j = \text{base}^{-j/d}$ is a predefined frequency, with $d$ being the hidden dimension size.
Then we have:
\[ (\mathbf{q}'_m)^H \mathbf{k}'_n = \sum_{j=1}^{d} \overline{\mathbf{q}}_{m,j} \mathbf{k}_{n,j} e^{i (n-m) \theta_j} \]
This result shows that the attention score is modulated by a relative phase shift $e^{i(n-m)\theta_j}$ that depends solely on the position difference $n-m$. The detailed derivation is provided in Appendix~\ref{justification:rope_derivation}.

\subsubsection{Layer Normalization.} The RMSNorm is applied to real and imaginary components of activations, respectively.

\subsection{\qscheme{} for Complex-Valued Weight}
\label{subsec:weight_quant}

The core of \mname{} lies in its quantization scheme for complex-valued weights, simulated during Quantization-Aware Training (QAT). We propose \qscheme{}, a deterministic method that maps each full-precision complex weight to one of the fourth roots of unity \(\{\pm1, \pm i\}\) based on its phase in the complex plane.
Each complex weight \(w = w_\text{re} + i w_\text{im}\) is first projected to a codeword using a phase-based mapping:
\[
\mathcal{P}(w) = i^{\left\lfloor \frac{2 \theta}{\pi} + \frac{1}{2} \right\rfloor}, \quad \theta = \text{Arg}(w) \in [-\pi, \pi].
\]
Letting 
\(w_b = \mathcal{P}(w) = w_{b,\text{re}} + i w_{b,\text{im}}\), we then compute the scaling factors \(\gamma_\text{re}\) and \(\gamma_\text{im}\) after this projection, using only the entries that are mapped to the corresponding codewords:
\[
\gamma_\text{re} = \frac{1}{\mathbb{E}\!\left[\,|\mathbf{W}_\text{re}| \;\middle|\; \mathcal{P}(\mathbf{W}) \in \{\pm 1\}\right]},
\]
\[
\gamma_\text{im} = \frac{1}{\mathbb{E}\!\left[\,|\mathbf{W}_\text{im}| \;\middle|\; \mathcal{P}(\mathbf{W}) \in \{\pm i\}\right]} .
\]
These factors normalize the respective components using only those entries whose phase-based projection falls into \(\{\pm1\}\) for \(\gamma_\text{re}\) or \(\{\pm i\}\) for \(\gamma_\text{im}\). 
Finally, the quantized value is dequantized as
\[
w_q = \frac{w_{b,\text{re}}}{\gamma_\text{re}} + i \cdot \frac{w_{b,\text{im}}}{\gamma_\text{im}} .
\]
During the forward pass, the full-precision weights are replaced with \(w_q\). In the backward pass, gradients are propagated through the original weights using the Straight-Through Estimator (STE) to handle the non-differentiable quantization step.
An illustration of \qscheme{} is provided in the left panel of Figure~\ref{fig:complexquant}.

\subsection{Complex-Valued Activation Quantization}
\label{subsec:act_quant}
We adopt a symmetric per-token INT8 quantization scheme that processes the real ($\mathbf{x}_\text{re}$) and imaginary ($\mathbf{x}_\text{im}$) components of the activation $\mathbf{x}$ independently. For each component, a dynamic scaling factor is computed based on the maximum absolute value within the token’s feature vector. Specifically, the scaling factor for the real part is:
\[
s_\text{re} = \frac{127}{\max(|\mathbf{x}_\text{re}|)}
\]
The quantization function, which emulates quantization effects during training, is defined as:
\[
\mathcal{Q}(\mathbf{x}, s) = \frac{1}{s} \cdot \text{round}\left(\text{clamp}(s \cdot \mathbf{x}, -128, 127)\right)
\]
The quantized activation is then reconstructed as:
$
\mathbf{x}_q = \mathcal{Q}(\mathbf{x}_\text{re}, s_\text{re}) + i \cdot \mathcal{Q}(\mathbf{x}_\text{im}, s_\text{im})
$
where $s_\text{im}$ is computed analogously. This per-token, component-wise quantization adapts to the varying dynamic range across tokens, yielding higher numerical precision than static per-tensor methods.

\subsection{Computational Complexity Analysis}
\label{subsec:complexity}
A key advantage of our approach is the enhancement of representational capacity without increasing computational overhead.

\subsubsection{Storage Cost.} The storage requirement for our model's weights is identical to that of the 1.58-bit BitNet. Each complex weight is stored using 2 bits to represent one of the four states $\{\pm1, \pm i\}$, thereby achieving maximum storage efficiency for a 2-bit system. The activations, quantized to INT8 for both real and imaginary parts, also follow standard low-precision data formats.

\subsubsection{Operational Cost.}
While a generic complex multiplication $(a+ib)(c+id) = (ac-bd) + i(ad+bc)$ requires four real multiplications and two real additions, the computation in \mname{} is substantially more efficient. Our quantized weights belong to a specific set that eliminates the need for multiplication. Let $w_q$ be a quantized weight and $x_q = x_\text{re} + ix_\text{im}$ be a quantized activation. The product $\overline{x}_q \cdot w_q$ results in one of four outcomes, as summarized in Table \ref{tab:ops}.

\begin{table}[ht]
\centering
\begin{tabular}{ccc}
\toprule
\textbf{Activation}&\textbf{Weight ($w_q$)} & \textbf{Result ($\overline{x}_q\cdot w_q$)} \\
\midrule
\multirow{4}{*}{$x_q = x_\text{re} + ix_\text{im}$} & $+1$ & $x_\text{re} - i x_\text{im}$ \\
&$-1$ & $-x_\text{re} + i x_\text{im}$ \\
&$+i$ & $x_\text{im} + i x_\text{re}$ \\
&$-i$ & $-x_\text{im} - i x_\text{re}$ \\
\bottomrule
\end{tabular}
\caption{Multiplication-free operations for the \clinear{} layer. $\overline{x}_q$ is the conjugate of the quantized activation.}
\label{tab:ops}
\end{table}
Crucially, all four operations are free of multiplications and can be implemented using additions, subtractions, and component swapping. This places the computational overhead of \mname{} in the same class as BitNet, dominated by additions. The matrix multiplication in our \clinear{} layer with \qscheme{} is thus transformed from a multiplication-intensive operation into an addition-intensive one.

\subsubsection{Inference Optimization with Look-Up Tables (LUTs).} The discrete nature of both our quantized weights (2-bit) and activations (INT8) makes the computation in \mname{} highly amenable to further optimization using look-up tables (LUTs), particularly for CPU inference. Following a similar principle to optimized kernels like `BitNet.cpp', the inner loop of the matrix multiplication can be significantly accelerated. For instance, a group of four 2-bit complex weights can be combined to form an 8-bit index ($4^4 = 256$ states). A LUT can be pre-computed to store the 256 possible outcomes of multiplying these four weights with a corresponding vector of four INT8 complex activations. During inference, the computation is transformed into fetching the pre-computed complex result from the LUT based on the weight configuration and accumulating it.

%% file: body/4eval.tex
\section{Experiments}
We conduct a comprehensive empirical evaluation of \mname{} to validate its effectiveness. 
We aim to answer the following key research questions: 
\begin{itemize}
    \item \textbf{RQ1 (Performance):} Does fully utilizing the 2-bit complex-valued quantization space lead to improved language modeling and downstream task performance compared to existing ternary quantization schemes? 
    \item \textbf{RQ2 (Ablation Insights):} How do complex-valued components of the \mname{} architecture, such as attention mechanism and LM head design, affect the performance? 
    \item \textbf{RQ3 (Quantization Dynamics):} Does the proposed quantization scheme make effective use of the full codebook $\{\pm1, \pm i\}$, and how do layer-wise $\ell_2$ norms behave under complex-valued quantization?
\end{itemize}

\subsection{Experimental Setup}
We outline the setup for our empirical evaluation in this section. We first list the models and baselines, followed by the evaluation protocol, and finally the implementation details.

\paragraph{Models and Baselines.}
We evaluate our proposed model, \mname{}, at 700M and 1.3B parameter scales. To provide a comprehensive performance context, we compare it against three primary baselines, each serving a distinct purpose:
\begin{itemize}
    \item \textbf{Full-Precision \mname{}}: Our own complex-valued architecture trained in full BF16 precision without quantizaiton. This model serves as the direct upper-bound for our approach, allowing us to isolate and measure the performance impact of quantization itself.
    \item \textbf{FP16 LLaMA}: A standard full-precision LLaMA model, acting as a widely-accepted performance benchmark for traditional real-valued Transformers.
    \item \textbf{BitNet b1.58}: A 1.58-bit LLM that serves as our main low-bit competitor. We compare against both officially reported results and our reproduction based on publicly available code\footnote{https://huggingface.co/1bitLLM} for a comprehensive comparison. Due to computational constraints, we only reproduce the 700M variant.
\end{itemize}

\paragraph{Evaluation Protocol.}
To comprehensively assess model capabilities, our evaluation is twofold:
\begin{itemize}
    \item \textbf{Language Modeling:} We measure perplexity (PPL) on the validation sets of WikiText2~\cite{merity2016pointer_wikitext2} and C4~\cite{raffel2020exploring_c4}. Lower PPL indicates superior language modeling ability.
    \item \textbf{Downstream Tasks:} We evaluate zero-shot performance on a suite of common sense reasoning tasks using the \texttt{lm-eval-harness} framework~\cite{eval-harness}. The tasks include ARC-Easy~\cite{yadav2019quick}, ARC-Challenge~\cite{yadav2019quick}, Hellaswag~\cite{zellers2019hellaswag}, Winogrande~\cite{sakaguchi2021winogrande}, and PIQA~\cite{bisk2020piqa}.
\end{itemize}

\paragraph{Implementation Details.}
All models are trained from scratch under a unified setting to ensure fair comparison. We use a 100B-token corpus randomly sampled from the RedPajama-V1 dataset~\cite{weber2024redpajama}, tokenized with the LLaMA-Tokenizer\footnote{https://huggingface.co/meta-llama/Llama-2-7b-hf}. Our models are trained using the AdamW optimizer~\cite{loshchilov2017adamw} with a two-stage linear learning rate decay schedule.
We divide the training process into two stages at the 50\% mark.
The first stage adopts standard linear learning rate scheduling with a higher peak learning rate, as 2-bit LLMs exhibit greater training stability than their full-precision counterparts. During this stage, the weight decay is set to 0.1. In the second stage, we apply a decayed learning rate schedule with a lower peak value, and the weight decay is reduced to 0.
For training efficiency and stability, we employ data parallelism and a BF16 mixed-precision strategy, where we train our \mname{} model in BF16 precision but accumulate gradients in FP32 precision. The training was conducted on a cluster of 32 NVIDIA H800 GPUs, leveraging HuggingFace Accelerate with the DeepSpeed (ZeRO Stage 1) backend. The key hyperparameters used for training \mname{} are summarized in Table \ref{tab:hyperparams}.

\begin{table}[ht]
\centering
\begin{tabular}{lc}
\toprule
\textbf{Hyperparameter} & \textbf{Value} \\
\midrule
Learning Rate (700M) &  $1.5\times 10^{-3}\rightarrow 1.0\times 10^{-3}$ \\
Learning Rate (1.3B) &  $1.2\times 10^{-3}\rightarrow 0.8\times 10^{-3}$ \\
LR Schedule & Two-stage linear \\
Warmup Steps & 375 \\
Adam $\beta_1$ & 0.9 \\
Adam $\beta_2$ & 0.95 \\
Weight Decay & 0.1$\rightarrow$ 0.0 \\
Gradient Clipping & 1.0 \\
Batch Size & 512\\
Sequence Length & 2048 \\
Mixed Precision & BF16 \\
\bottomrule
\end{tabular}
\caption{Key training hyperparameters for \mname{}.}
\label{tab:hyperparams}
\end{table}

\subsection{Main Results}
\label{subsec:main_results}
\begin{figure}[ht]
    \centering
    \begin{subfigure}[t]{0.45\linewidth}
        \centering
        \includegraphics[width=\linewidth]{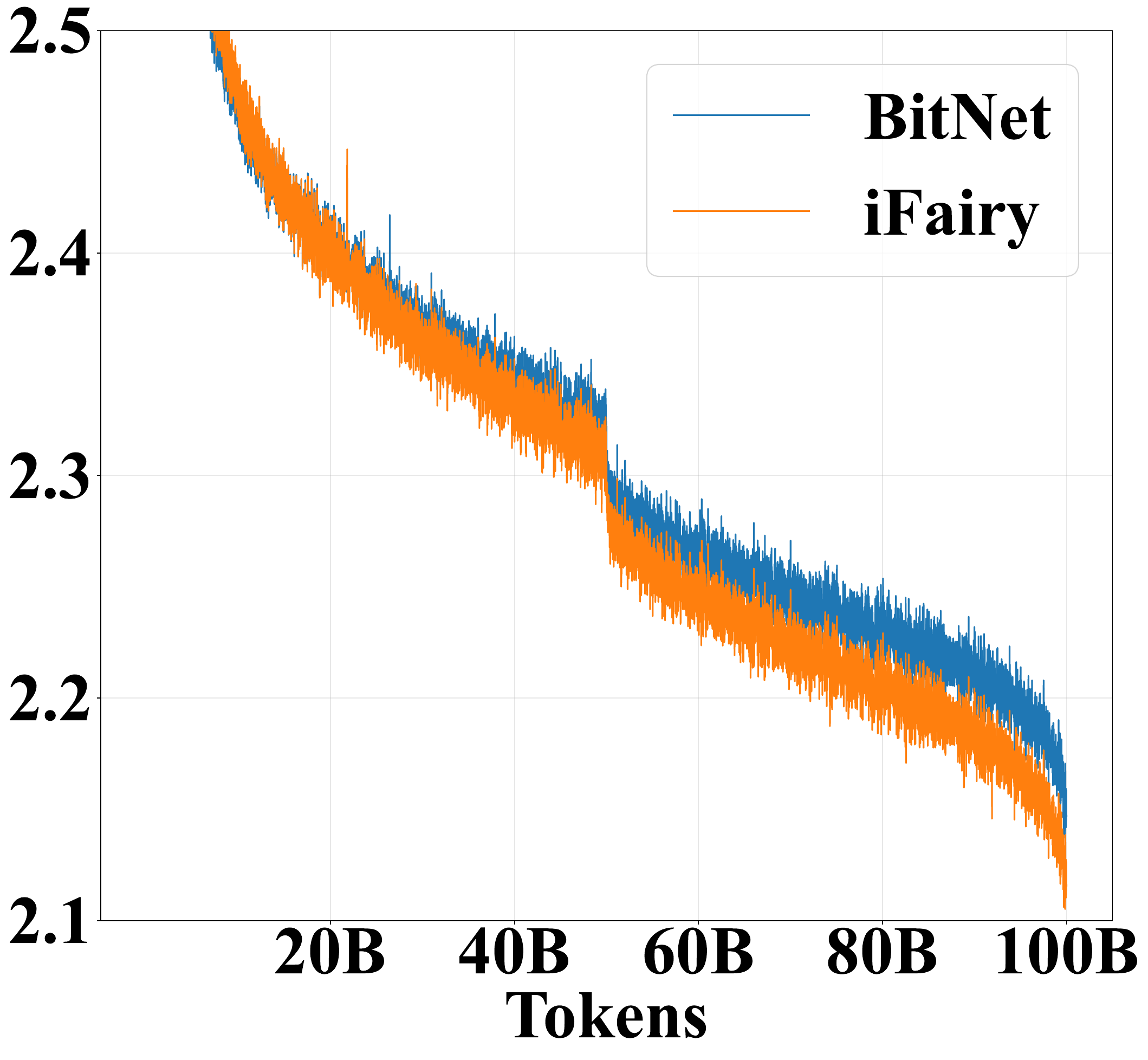}
        \caption{Training loss curve of \mname{} and BitNet b1.58.}
        \label{fig:training_loss}
    \end{subfigure}
    \hfill
    \begin{subfigure}[t]{0.45\linewidth}
        \centering
        \includegraphics[width=\linewidth]{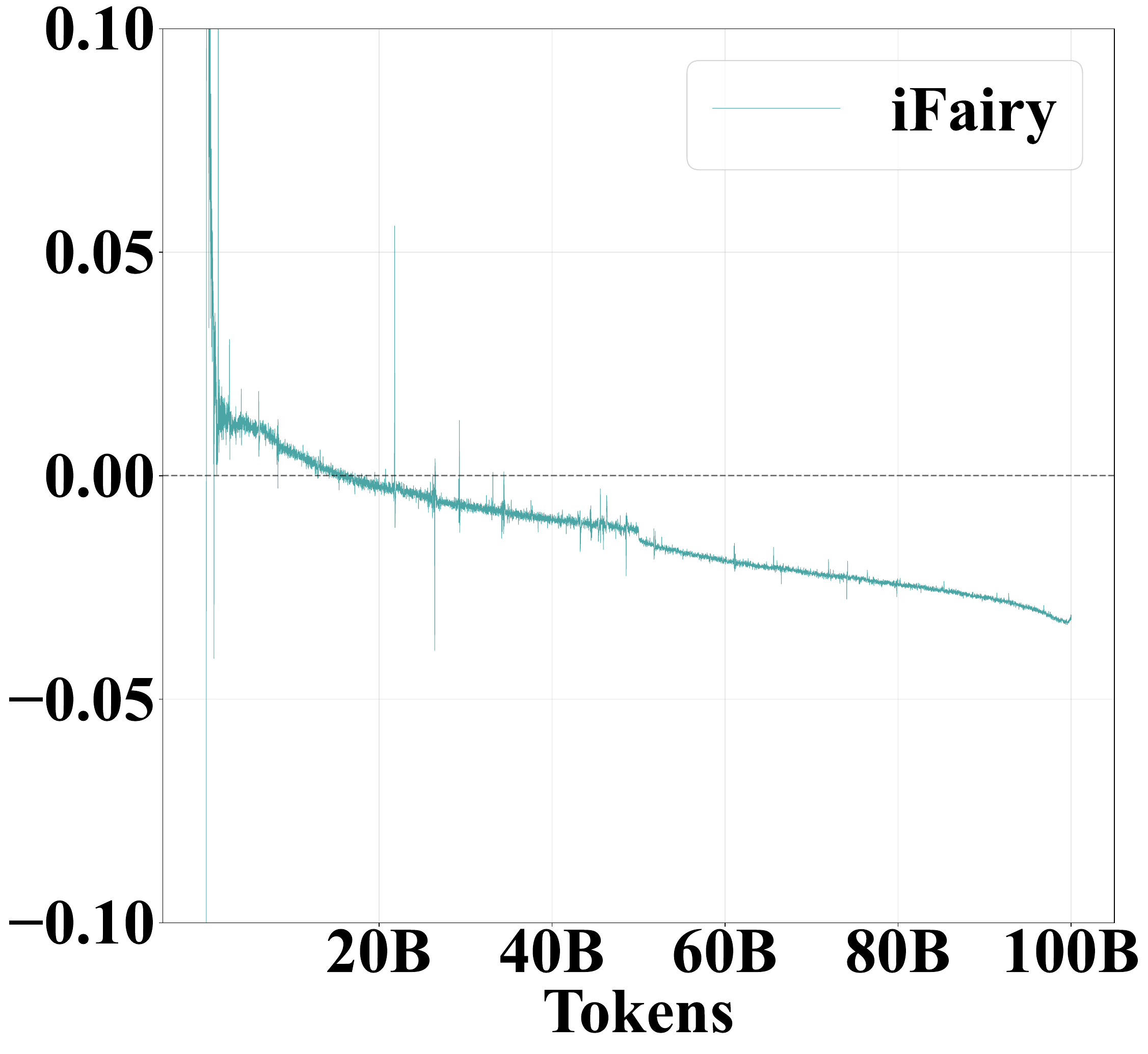}
        \caption{Training loss difference between \mname{} and BitNet b1.58.}
        \label{fig:loss_diff}
    \end{subfigure}
    \caption{Training loss comparision between \mname{} and BitNet b1.58.}
    \label{fig:loss_comparison}
\end{figure}

In this section, we present the core empirical results to answer \textbf{RQ1}. We demonstrate this through a top-down analysis, starting from training dynamics, followed by language modeling perplexity, and finally, downstream task performance.
\paragraph{Training Dynamics and Convergence.}
We begin by examining the training loss, a fundamental indicator of a model's ability to learn from data. Figure~\ref{fig:loss_comparison} compares the training loss curves of \mname{} and BitNet b1.58. \mname{} consistently achieves a lower training loss throughout the training process, indicating that our complex-valued quantization scheme enables more effective optimization and a better fit to the training data. This superior convergence behavior lays the foundation for its strong performance on evaluation benchmarks.

\paragraph{Language Modeling Performance.}
This improved training dynamic translates directly into superior language modeling capabilities. Table~\ref{tab:ppl_grouped} presents the perplexity (PPL) scores on the WikiText2 and C4 validation sets. 
Our method, \mname{}, consistently outperforms both the reproduced and reported versions of BitNet b1.58 across model sizes. 
At the 700M scale, \mname{} achieves an average PPL of 11.13, improving upon BitNet b1.58's 11.51 (reproduced) and 12.87 (reported). 
At the 1.3B scale, \mname{} achieves an average PPL of 10.14, significantly lower than the 11.29 reported for BitNet b1.58. These results confirm that our complex-valued, 2-bit quantization framework enhances model expressiveness under extreme compression.

\begin{table}[ht]
\centering
\setlength{\tabcolsep}{4pt}
\renewcommand{\arraystretch}{1.1}
\begin{tabular}{llcccc}
\toprule
\textbf{Size} & \textbf{Model} & \textbf{Quant} & \textbf{Wiki2$\downarrow$} & \textbf{C4$\downarrow$} & \textbf{Avg$\downarrow$} \\
\midrule
\multirow{5}{*}{\textbf{700M}}
& FP16 LLaMA & No & -- & -- & 12.33 \\
& \text{\mname{}}$^{\circ}$(Ours) & No & \textbf{9.41} & \textbf{10.75} & \textbf{10.08} \\
\cmidrule(l){2-6}
& BitNet b1.58$^{\star}$   & Yes & --    & --    & 12.87 \\
& BitNet b1.58$^{\dagger}$ & Yes & 10.81 & 12.21 & 11.51 \\
& \textbf{\mname{}} (Ours)  & Yes & \textbf{10.45} & \textbf{11.81} & \textbf{11.13} \\
\midrule
\multirow{4}{*}{\textbf{1.3B}}
& FP16 LLaMA & No & -- & -- & 11.25 \\
& \text{\mname{}}$^{\circ}$(Ours) & No & \textbf{8.72} & \textbf{9.95} & \textbf{9.34} \\
\cmidrule(l){2-6}
& BitNet b1.58$^{\star}$ & Yes & -- & -- & 11.29 \\
& \textbf{\mname{}} (Ours) & Yes & \textbf{9.35} & \textbf{10.94} & \textbf{10.14} \\
\bottomrule
\end{tabular}
\caption{
Perplexity on WikiText2 and C4 validation sets (lower is better). $^{\star}$ refers to the reported version in prior work~\cite{ma2024bitnetb1.58}, $^{\dagger}$ our reproduced version, and $^{\circ}$ the full-precision \mname{}.
}
\label{tab:ppl_grouped}
\end{table}

\paragraph{Downstream Task Performance.}
To assess how well these improvements generalize beyond the pre-training objective, we evaluate the models on a suite of zero-shot common sense downstream tasks. The results, summarized in Table~\ref{tab:downstream_grouped}, highlight the strong generalization capacity of \mname{}. Remarkably, our 1.3B \mname{} model achieves an average accuracy of 46.52, not only exceeding the BitNet baseline but also slightly outperforming the FP16 LLaMA model (46.21). This finding underscores that the rich representations learned via our 2-bit complex quantization are highly effective and transferable to diverse downstream applications.

\begin{table*}[th]
\centering
\setlength{\tabcolsep}{5.5pt}
\renewcommand{\arraystretch}{1.1}
\begin{tabular}{llccccccccc}
\toprule
\textbf{Model Size} & \textbf{Model} & \textbf{Quant} & \textbf{ARCe$\uparrow$} & \textbf{ARCc$\uparrow$} & \textbf{HS$\uparrow$} & \textbf{BQ$\uparrow$} & \textbf{OQ$\uparrow$} & \textbf{PQ$\uparrow$} & \textbf{WGe$\uparrow$} & \textbf{Avg$\uparrow$} \\
\midrule
\multirow{5}{*}{\textbf{700M}} 
& FP16 LLaMA & No & 54.70 & 23.00 & 37.00 & 60.00 & 20.20 & 68.90 & \textbf{54.80} & 45.51 \\
& \text{\mname{}}$^{\circ}$(Ours) & No & \textbf{55.68} & \textbf{24.06} & \textbf{37.79} & \textbf{60.46} & \textbf{20.60} & \textbf{70.18} & 54.46 & \textbf{46.18} \\
\cmidrule(l){2-11}
& BitNet b1.58$^{\star}$ & Yes & 51.80 & 21.40 & 35.10 & 58.20 & 20.00 & \textbf{68.10} & \textbf{55.20} & 44.26 \\
& BitNet b1.58$^{\dagger}$ & Yes & 51.77 & 22.44 & 35.30 & \textbf{58.50} & 20.80 & 65.94 & 54.85 & 44.23 \\
& \text{\mname{}} (Ours) & Yes & \textbf{53.45} & \textbf{23.04} & \textbf{36.04} & 57.31 & \textbf{21.00} & 68.01 & 54.06 & \textbf{44.70} \\
\midrule
\multirow{4}{*}{\textbf{1.3B}}
& FP16 LLaMA & No & 56.90 & 23.50 & 38.50 & 59.10 & 21.60 & 70.00 & 53.90 & 46.21 \\
& \text{\mname{}}$^{\circ}$(Ours) & No & \textbf{58.96} & \textbf{25.77} & \textbf{40.29} & \textbf{60.92} & \textbf{23.20} & \textbf{71.44} & \textbf{57.06} & \textbf{48.23} \\
\cmidrule(l){2-11}
& BitNet b1.58$^{\star}$ & Yes & 54.90 & 24.20 & 37.70 & 56.70 & 19.60 & 68.80 & \textbf{55.80} & 45.39 \\
& \text{\mname{}} (Ours) & Yes & \textbf{56.65} & \textbf{24.66} & \textbf{38.69} & \textbf{59.60} & \textbf{22.20} & \textbf{69.80} & 54.06 & \textbf{46.52} \\
\bottomrule
\end{tabular}
\caption{Zero-shot accuracy on downstream tasks. $^{\star}$ refers to the reported version in prior work~\cite{ma2024bitnetb1.58}, $^{\dagger}$ our reproduced version, and $^{\circ}$ the full-precision \mname{}.}
\label{tab:downstream_grouped}
\end{table*}

\subsection{Ablation Studies}
To dissect the sources of \mname{}'s better performance and answer question \textbf{RQ2}, we conduct targeted ablation studies. We structure this analysis into two parts: first, we evaluate the inherent potential of a native complex-valued architecture, and second, we isolate the specific impact of our proposed quantization scheme, \qscheme{}.

\paragraph{Performance of Native Complex-Valued Architecture.} 
Before assessing our quantization scheme, it is crucial to establish the viability of a complex-valued Transformer as a strong architectural foundation. To this end, we compare our full-precision \mname{}, the model denoted as $\text{\mname{}}^{\circ}$ in Table~\ref{tab:ppl_grouped}, which is trained in BF16 without any quantization, against the standard FP16 LLaMA.

As shown in Table~\ref{tab:ppl_grouped} and Table~\ref{tab:downstream_grouped}, the native complex-valued architecture shows a distinct performance advantage. In language modeling, the full-precision \mname{} of 700M achieves a striking average PPL of 10.08, substantially outperforming the 12.33 of the FP16 LLaMA. This superiority in core modeling capability extends to downstream generalization; $\text{\mname{}}^{\circ}$ attains a higher average accuracy of 46.18 on downstream tasks, compared to 45.51 for its real-valued counterpart. These results confirm that a native complex-valued architecture inherently possesses greater representational power than a similar-scale real-valued architecture. This not only justifies our choice of architecture but also successfully breaks through the previous informational ceiling.

\begin{figure}[ht]
    \centering
    \begin{subfigure}[t]{0.45\linewidth}
        \centering
        \includegraphics[width=\linewidth]{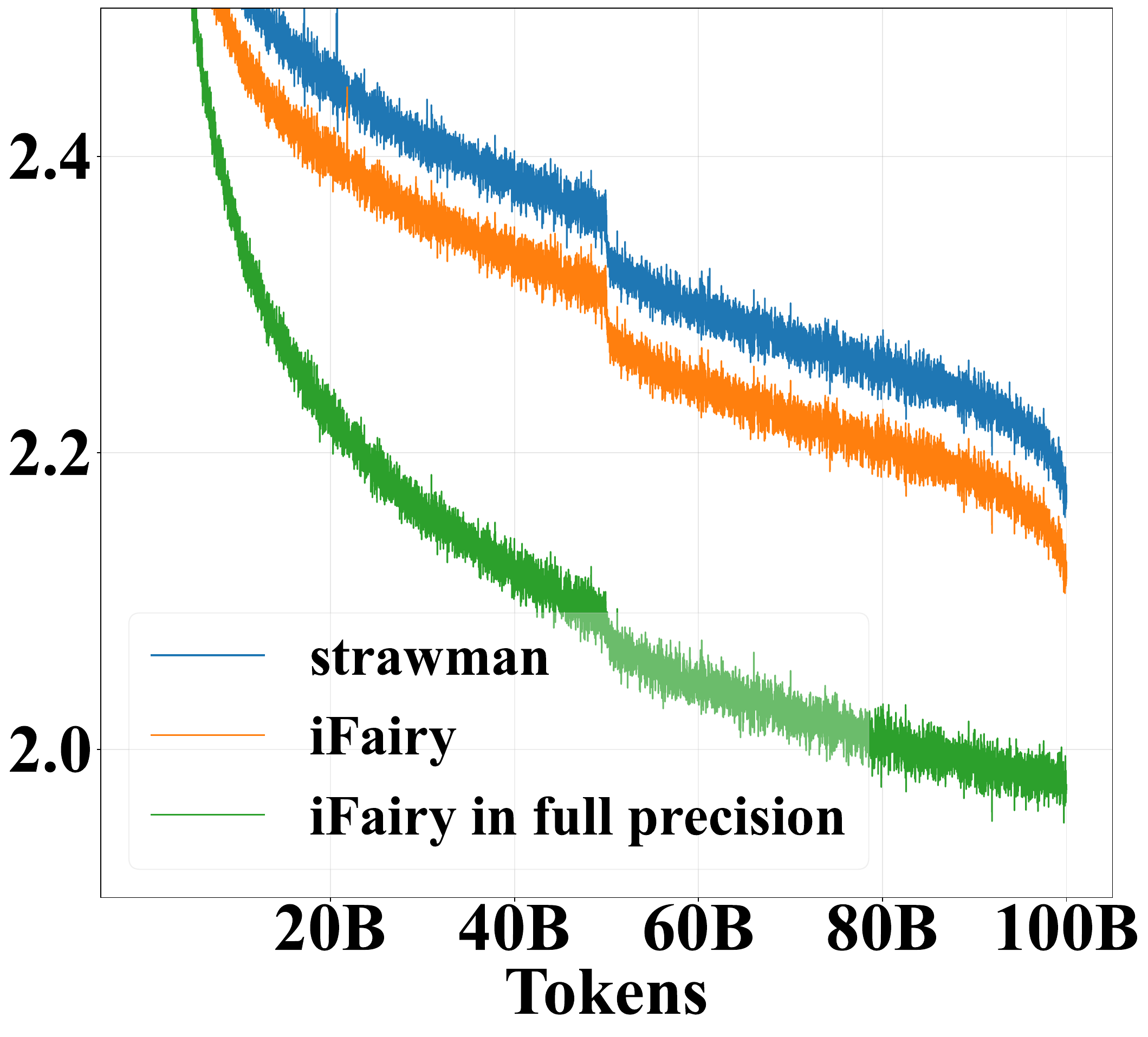}
        \caption{Training loss curve.}
        \label{fig:ablation_activation}
    \end{subfigure}
    \hfill
    \begin{subfigure}[t]{0.45\linewidth}
        \centering
        \includegraphics[width=\linewidth]{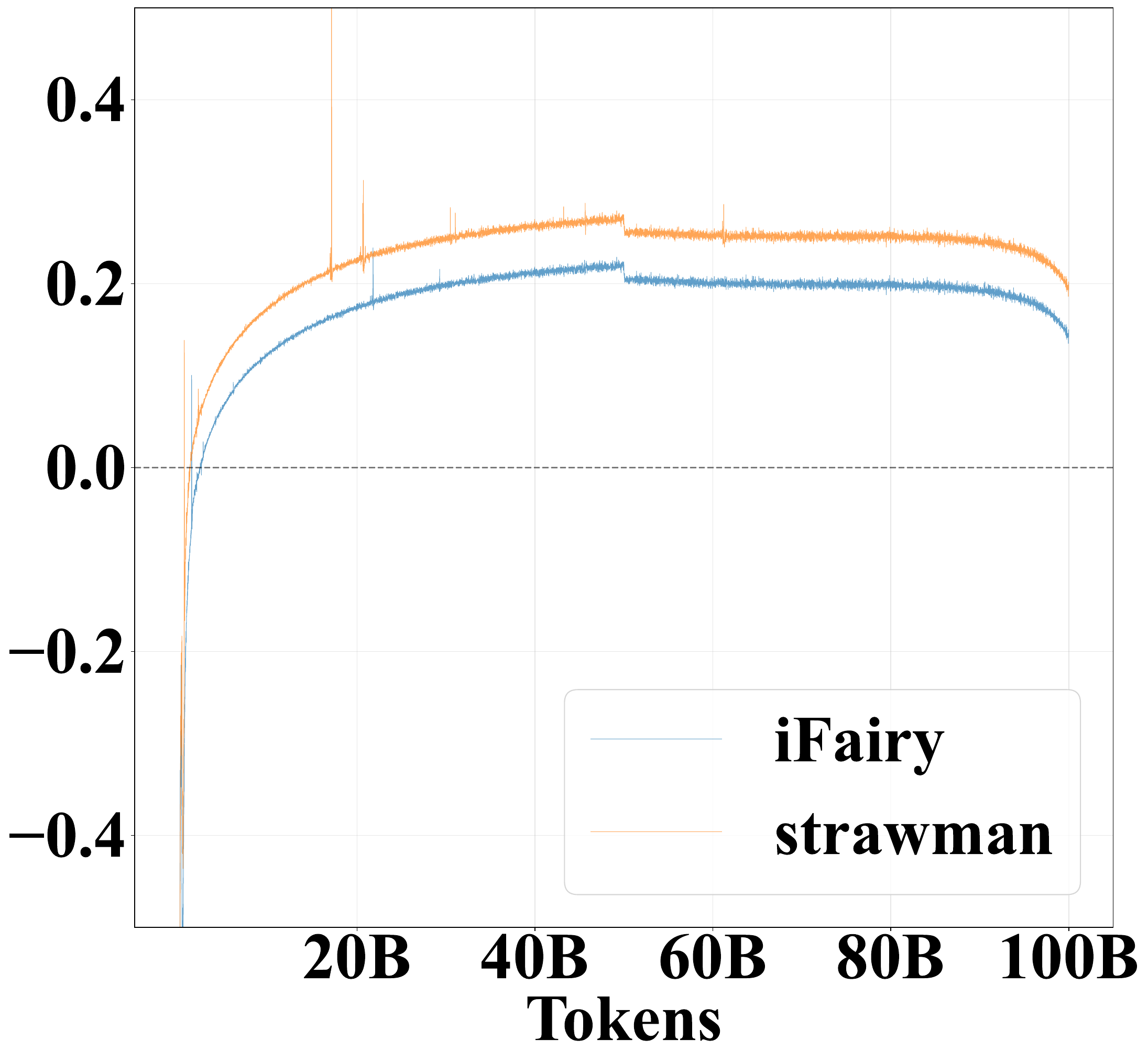}
        \caption{Training loss difference.}
        \label{fig:ablation_lmhead}
    \end{subfigure}
    \caption{Training loss comparison among \mname{}, full-precision \mname{} and the strawman solution with simple computational pattern. We use the full-precision \mname{} as the baseline of loss difference.}
    \label{fig:computational_pattern}
\end{figure}

\paragraph{Impact of Computational Pattern.} 
A critical design choice within a complex-valued architecture is the specific computational pattern used to handle interactions between complex states.
A strawman solution is to compute the attention score from the dot product of only the real parts of the query and key vectors. Similarly, the LM head projection uses only the real part of the final hidden state. 
As shown in the training loss comparison in Figure~\ref{fig:computational_pattern}, the choice of pattern has a profound impact on training dynamics. The persistent gap between the two patterns demonstrates that our proposed computational pattern better leverages the expressive power of the complex domain by enabling the model to jointly process both real and imaginary parts throughout the network. This refined handling of information is a key contributor to the superior final performance of \mname{}.

\subsection{Analysis of Complex-Valued Quantized Representations}
\label{subsec_quant_analysis}
To verify that the proposed quantization scheme fully utilizes the 2-bit space, we analyze the intrinsic properties of our quantization scheme to answer our final research question \textbf{RQ3}. 
We examine this through three complementary perspectives: (1) the distribution of quantized weights across the 2-bit complex codebook, (2) the behavior of layer-wise weight norms, and (3) the distribution of the token embedding layer and the LM head.

\subsubsection{Distribution of Quantized Model Weights.} 
\begin{figure}[ht]
    \centering
    \begin{subfigure}[t]{0.45\linewidth}
        \centering
        \includegraphics[width=\linewidth]{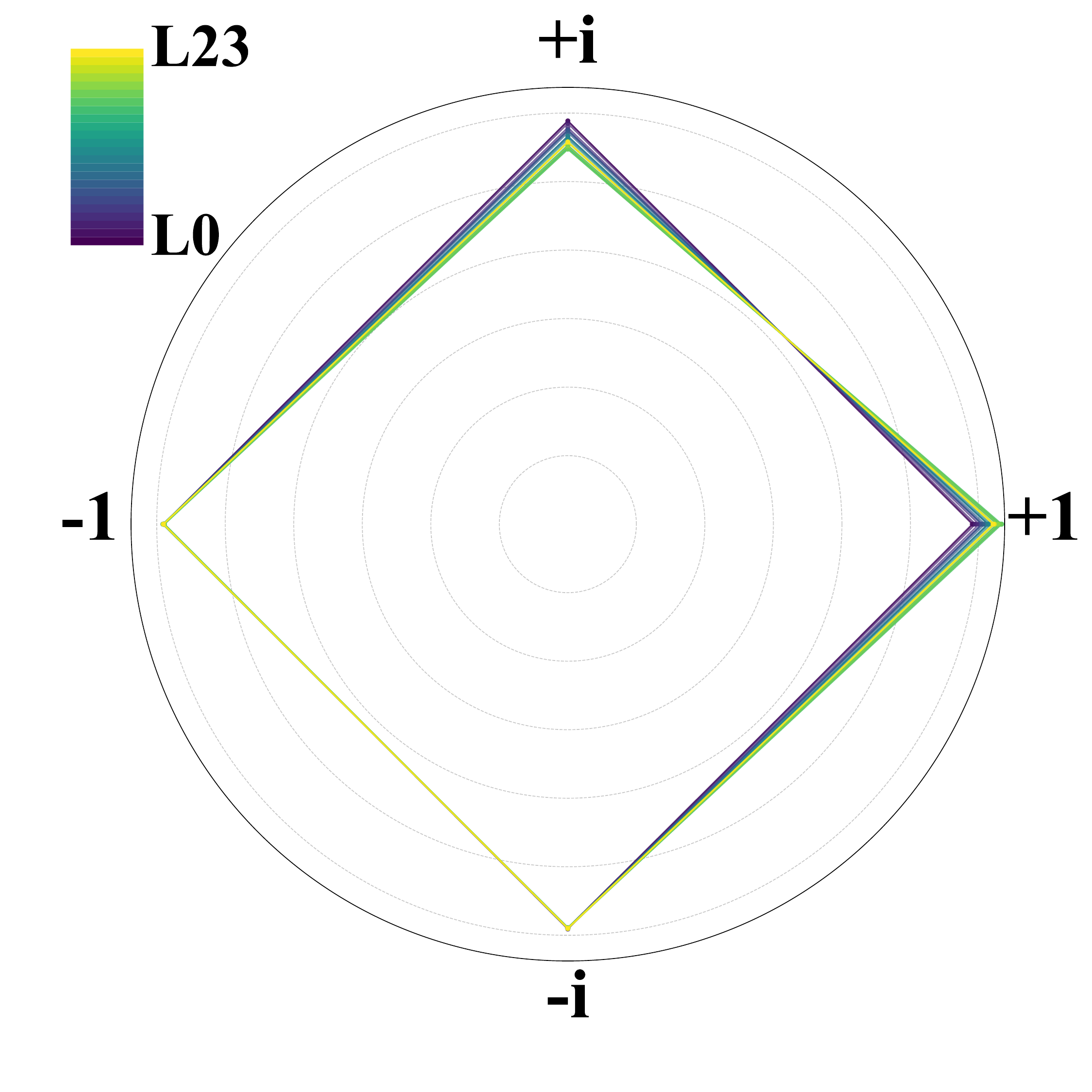}
        \caption{Empirical distribution of quantized weights in $\mathbf{W}_\mathbf{K}$.}
        \label{fig:value_distribution1}
    \end{subfigure}
    \hfill
    \begin{subfigure}[t]{0.45\linewidth}
        \centering
        \includegraphics[width=\linewidth]{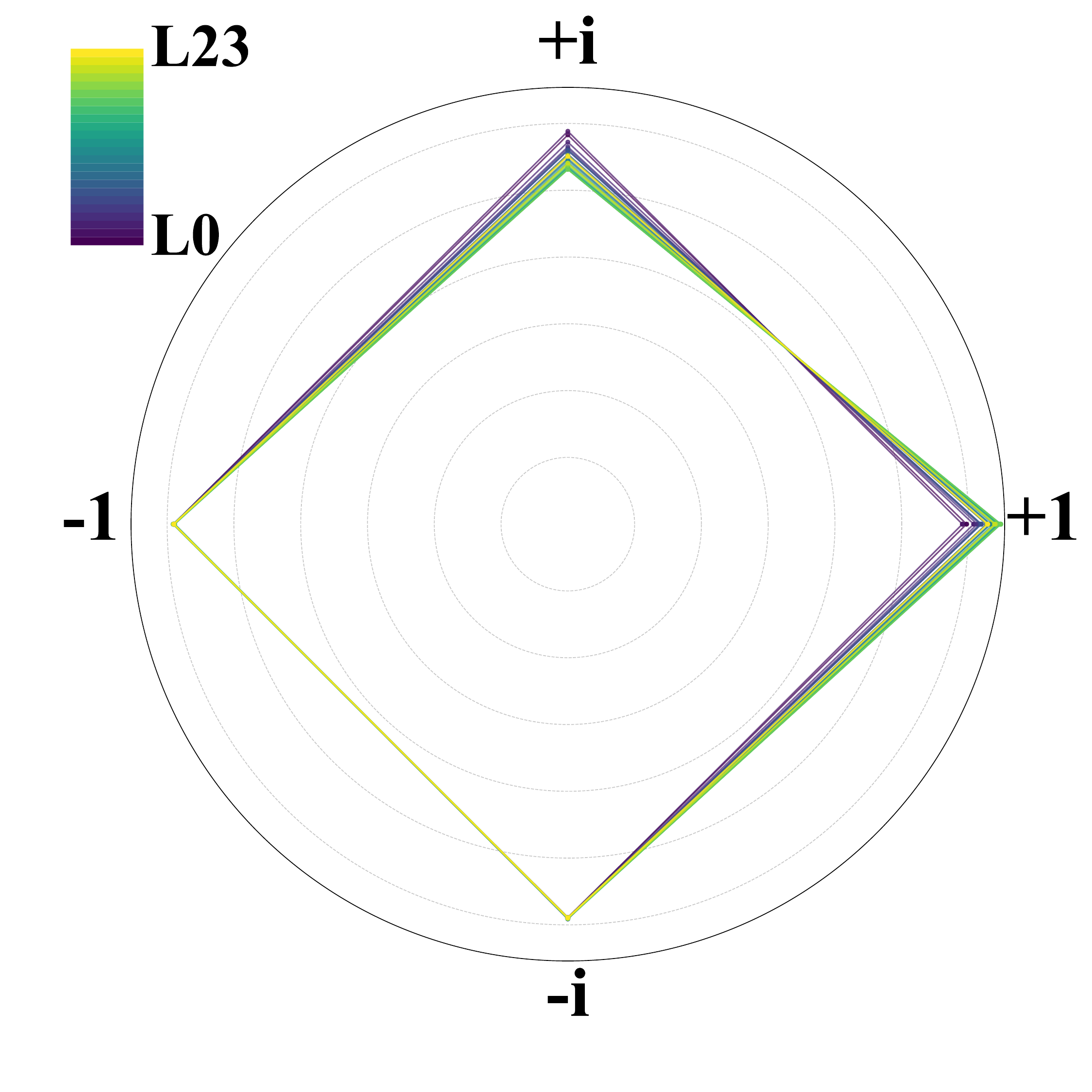}
        \caption{Empirical distribution of quantized weights in $\mathbf{W}_\mathbf{O}$.}
        \label{fig:value_distribution2}
    \end{subfigure}
    \caption{Quantization statistics of weight values in \mname{}.}
    \label{fig:value_distribution}
\end{figure}

A key indicator of an effective multi-bit quantization scheme is its ability to leverage the entire available representational space. A poorly designed scheme might lead to representational collapse, where the model predominantly uses only a subset of the available values. We measured the empirical distribution of its quantized weights across the four complex values $\{\pm1, \pm i\}$ to demonstrate codebook utilization of \mname{}.
As shown in Figure~\ref{fig:value_distribution}, the distribution is remarkably balanced. This near-uniformity confirms that the model effectively learns to exploit the full expressive power of the 2-bit complex codebook. Every quantum state is actively used, providing the model with the rich representational capacity that underpins its strong performance.
For completeness, we include the weight distributions of other modules in Appendix~\ref{app:weight_distribution}, which demonstrate similarly uniform usage patterns.

\subsubsection{Layer-wise Norms of Quantized Weights.}
\begin{figure}[ht]
    \centering
    \begin{subfigure}[t]{0.45\linewidth}
        \centering
        \includegraphics[width=\linewidth]{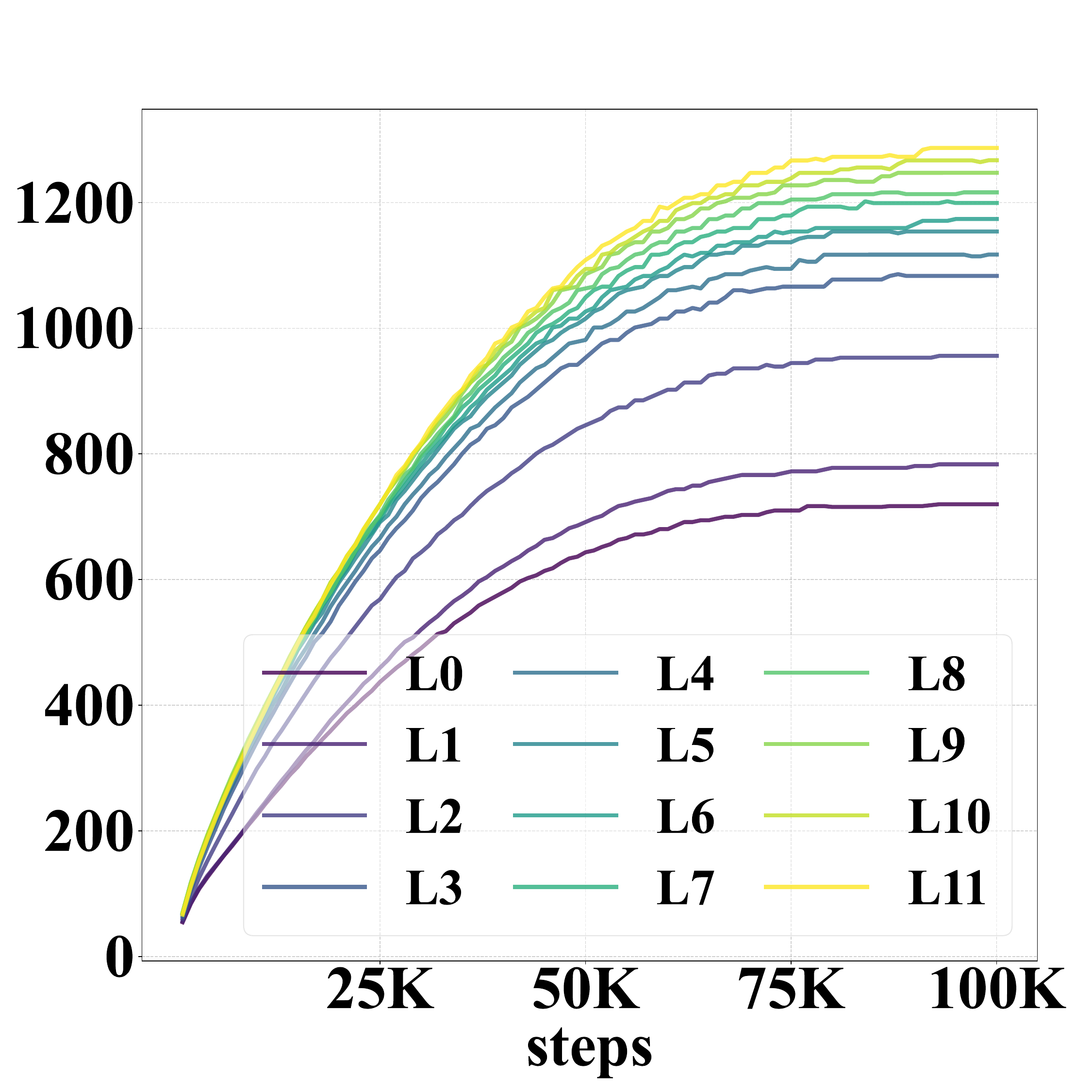}
        \caption{Down projection $\ell_2$ norm across layer 0 to layer 11.}
        \label{fig:l2_norm_a}
    \end{subfigure}
    \hfill
    \begin{subfigure}[t]{0.45\linewidth}
        \centering
        \includegraphics[width=\linewidth]{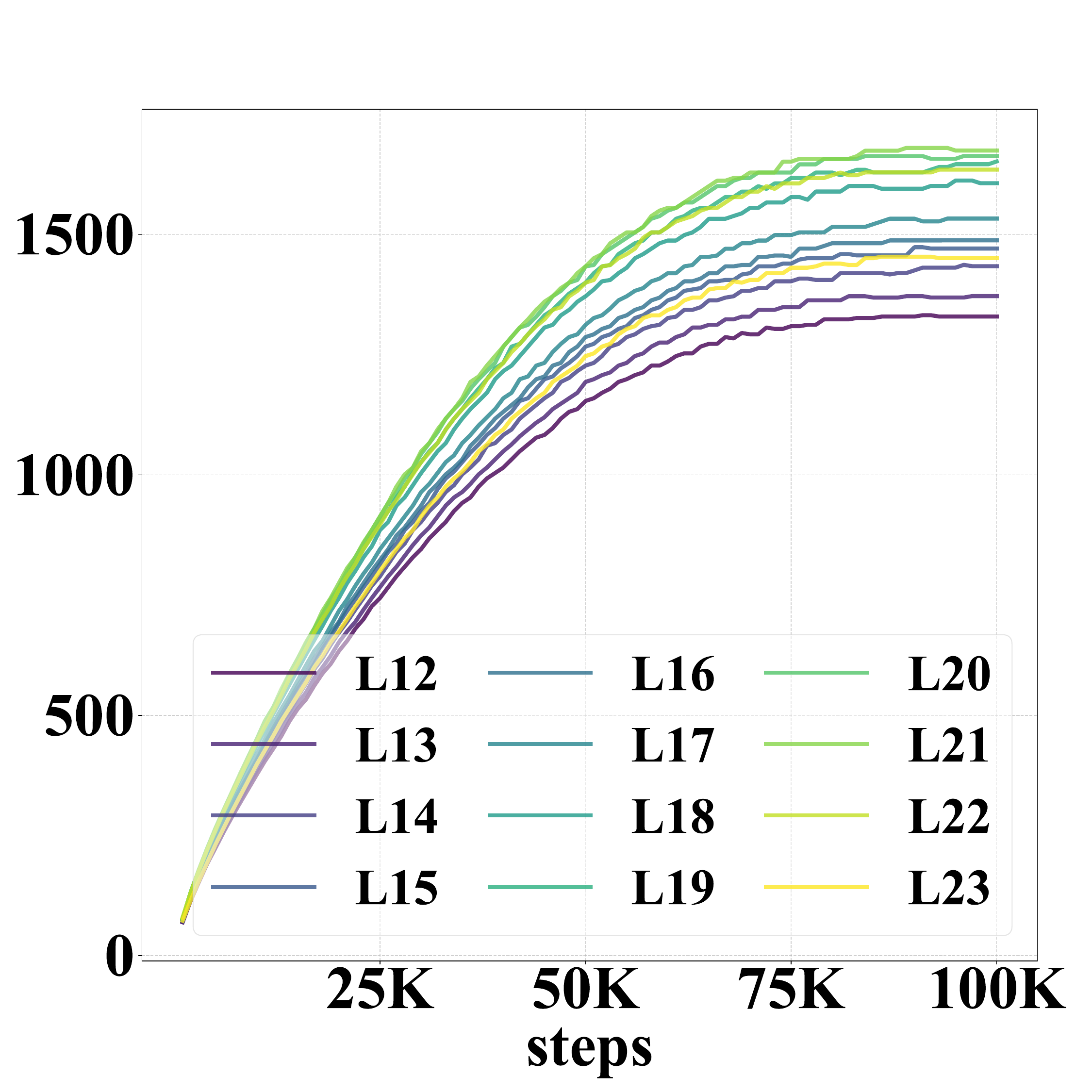}
        \caption{Down projection $\ell_2$ norm across layer 12 to layer 23.}
        \label{fig:l2_norm_b}
    \end{subfigure}
    \caption{Layer-wise $\ell_2$ norm of complex-valued quantized weights in \mname{}. Stable norm patterns suggest good magnitude preservation and strong generalization capacity.}
    \label{fig:l2_norm}
\end{figure}

Beyond utilizing the codebook, a robust quantization method must maintain the structural integrity of the network. Unstable weight magnitudes across layers can hinder training and harm generalization. We therefore analyze the layer-wise $\ell_2$ norms of the quantized weights. 
Figure~\ref{fig:l2_norm} reveals that the norms remain exceptionally stable and well-distributed across all layers. This demonstrates that our method, including the separate scaling of real and imaginary components, successfully preserves the network's magnitude structure. Such stability is critical for maintaining healthy gradient flow throughout the deep network, preventing issues common in highly compressed models and contributing directly to the robust generalization we observed in Section~\ref{subsec:main_results}. For completeness, we provide the layer-wise norm statistics for all other weight matrices, in Appendix~\ref{app:l2norms}, which exhibit similar stability trends.

\subsubsection{Distribution of Embedding Layer and LM Head.}
\begin{figure}[ht]
    \centering
    \begin{subfigure}[t]{0.45\linewidth}
        \centering
        \includegraphics[width=\linewidth]{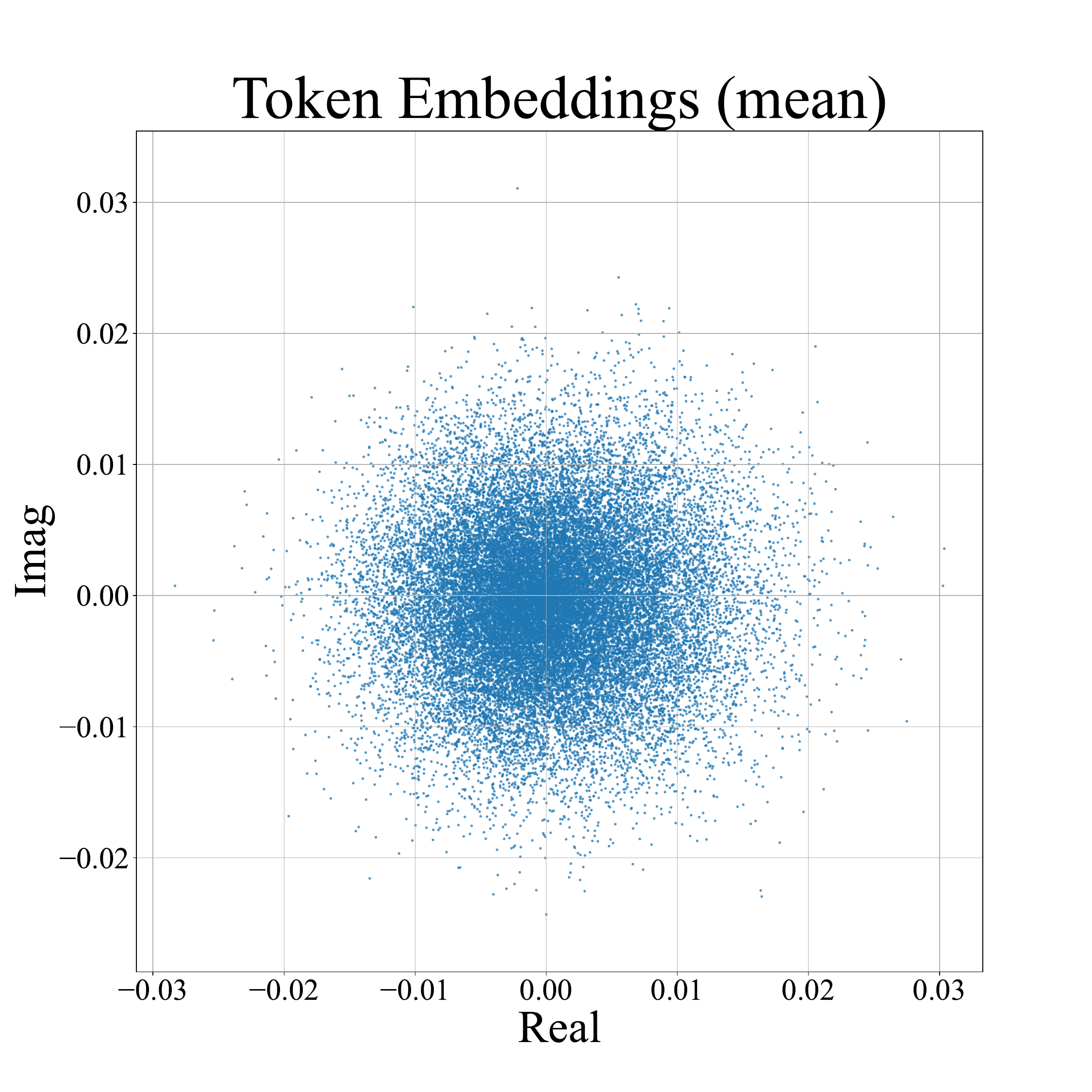}
        \caption{Mean-centered token embeddings plotted in the complex plane.}
        \label{fig:token_a}
    \end{subfigure}
    \hfill
    \begin{subfigure}[t]{0.45\linewidth}
        \centering
        \includegraphics[width=\linewidth]{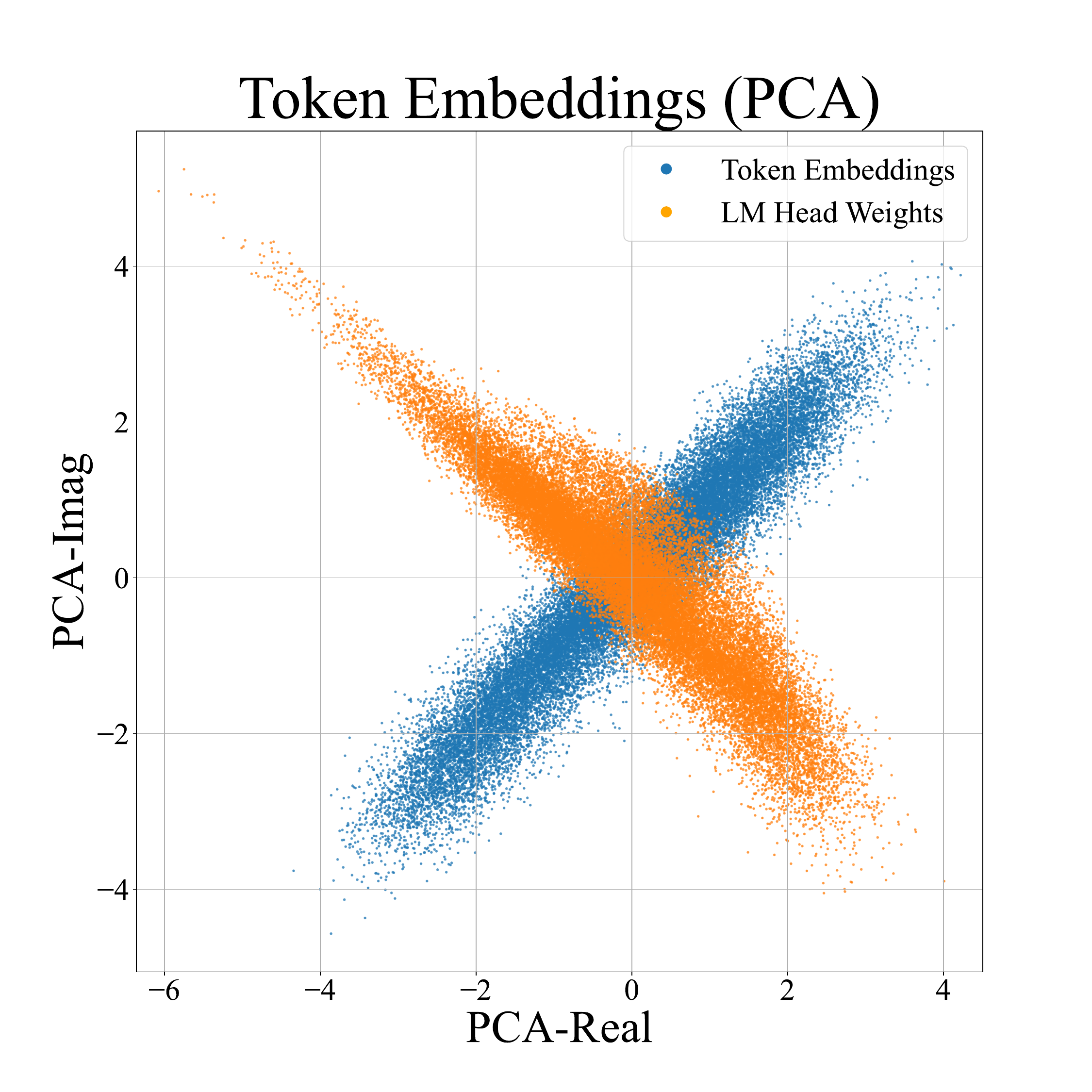}
        \caption{2D PCA projection of token embeddings and LM head weights.}
        \label{fig:token_b}
    \end{subfigure}
    \caption{Visualization of Complex-Valued Token Embeddings and LM Head Weights.}
    \label{fig:token}
\end{figure}

Figure~\ref{fig:token} visualizes the distribution of token embeddings and LM head weights in the complex plane. We plot the mean-centered token embeddings in their original complex form in Figure\ref{fig:token_a}. The points exhibit an approximately uniform distribution around the origin, indicating that both real and imaginary components are utilized in a balanced manner.
We apply principal component analysis (PCA) separately to the real and imaginary parts to project the embeddings and LM head weights into a 2D space in Figure\ref{fig:token_b}. The token embeddings (blue) and the LM head weights (orange) form well-separated but symmetric and coherent clusters, aligned along orthogonal directions. This structured distribution suggests that the complex-valued embedding space remains well-established and that the LM head learns to align with the token embedding.
Overall, the results show that our complex-valued architecture maintains a stable and expressive embedding space.

Taken together, these analyses provide strong evidence that our complex-valued quantization scheme is both expressive and stable. The full utilization of the 2-bit codebook ensures that the model leverages the complete representational capacity available in complex space. The stable layer-wise norm dynamics indicate that the quantized weights retain well-conditioned magnitudes across the network. The structural alignment between token embeddings and the LM head demonstrates that our model has a stable and uniform embedding space in the complex domain. 
These properties jointly underpin the robust performance of \mname{} under extreme bit-width constraints.

%% file: body/5conclusion.tex
\section{Conclusion}
We present \mname{}, the first 2-bit complex LLM with all parameters in $\{\pm1, \pm i\}$. 
We integrate complex-valued representations into the Transformer and quantize weights to the fourth roots of unity $\{\pm1, \pm i\}$ via the proposed \qscheme{}, \mname{} fully exploits the 2-bit space while preserving symmetry, efficiency, and hardware compatibility.
Experimental results demonstrate that \mname{} outperforms the accuracy ceiling of all existing quantization approaches under equivalent model sizes, in terms of perplexity and task accuracy. 

\paragraph{Limitations and Future Work.}
Several limitations remain. First, the optimal formulation of a complex-valued attention mechanism for language modeling is still underexplored. Second, our use of separate scaling factors for the real and imaginary components may not fully preserve the original magnitude structure of complex weights. Third, deploying \mname{} in practical systems demands careful hardware-aware design, as current CPU and GPU architectures are not optimized for complex-valued or multiplication-free computation.
Future work will focus on scaling \mname{} to larger model sizes, exploring unified or learned scaling strategies, and developing hardware accelerators tailored to complex-valued arithmetic. We also envision the design of more expressive, complex-native architectures that further enhance the benefits of complex quantization.

%% file: body/appendix.tex
\clearpage
\setcounter{secnumdepth}{2}  %
\appendix
\section{Pseudo-Code}
\label{appendix:code}

\begin{algorithm}[ht]
\caption{Forward Pass during QAT of \mname{}}
\label{alg:complexlinear}
\begin{algorithmic}[1]
\STATE \textbf{Input:} Full-precision complex weight $\mathbf{W}$, Full-precision complex activation $\mathbf{X}$
\STATE \textbf{Output:} Full-precision complex output $\mathbf{Y}$
\STATE
\STATE \COMMENT{1. Activation Quantization-Dequantization}
\STATE $s_{\text{re}} \gets 127 / \max(|\mathbf{X}_{\text{re}}|)$; \quad $s_{\text{im}} \gets 127 / \max(|\mathbf{X}_{\text{im}}|)$
\STATE $\mathbf{X}_{\text{int8,re}} \gets \text{round}(\text{clamp}(\mathbf{X}_{\text{re}} \cdot s_{\text{re}}, -128, 127))$
\STATE $\mathbf{X}_{\text{int8,im}} \gets \text{round}(\text{clamp}(\mathbf{X}_{\text{im}} \cdot s_{\text{im}}, -128, 127))$
\STATE $\mathbf{X}_{\text{q,re}} \gets \mathbf{X}_{\text{int8,re}} / s_{\text{re}}$
\STATE $\mathbf{X}_{\text{q,im}} \gets \mathbf{X}_{\text{int8,im}} / s_{\text{im}}$
\STATE $\mathbf{X}_{q} \gets \mathbf{X}_{\text{q,re}} + i\,\mathbf{X}_{\text{q,im}}$
\STATE
\STATE \COMMENT{2. Weight Quantization-Dequantization}
\STATE $\mathbf{W}_{\text{b}} \gets \mathcal{P}(\mathbf{W})$ \COMMENT{Quantized to $\{\pm1, \pm i\}$}
\STATE $\gamma_\text{re} \gets 1/\mathbb{E}\!\left[\,|\mathbf{W}_\text{re}| \;\middle|\; \mathcal{P}(\mathbf{W}) \in \{\pm 1\}\right]$

\STATE $\gamma_\text{im} \gets 1/\mathbb{E}\!\left[\,|\mathbf{W}_\text{im}| \;\middle|\; \mathcal{P}(\mathbf{W}) \in \{\pm i\}\right]$

\STATE $\mathbf{W}_{\text{q,re}} \gets \mathbf{W}_{\text{b,re}} / \gamma_{\text{re}}$
\STATE $\mathbf{W}_{\text{q,im}} \gets \mathbf{W}_{\text{b,im}} / \gamma_{\text{im}}$
\STATE $\mathbf{W}_{q} \gets \mathbf{W}_{\text{q,re}} + i\,\mathbf{W}_{\text{q,im}}$
\STATE
\STATE \COMMENT{3. Perform Complex Linear Operation}
\STATE $\mathbf{Y}' \gets \overline{\mathbf{X}_{q}} \cdot \mathbf{W}_{q}$ \COMMENT{Note the conjugate on $\mathbf{X}_{q}$}
\STATE
\STATE \COMMENT{4. Straight-Through Estimator (STE)}
\STATE \COMMENT{In backward pass, gradient flows to original $\mathbf{W}$}
\STATE $\mathbf{Y} \gets \mathbf{Y}'$
\STATE \textbf{return} $\mathbf{Y}$
\end{algorithmic}
\end{algorithm}

\section{Training Details}
\label{appendix:training}
\subsection{Hardware and Software Configuration}
All models were trained on a high-performance computing cluster equipped with 32 × NVIDIA H800 Tensor Core GPUs (80 GB HBM3 memory per GPU). 
Training was performed with DeepSpeed ZeRO Stage 1 for optimizer state sharding and memory efficiency, and bf16 mixed precision was used to reduce memory footprint and improve computational throughput while maintaining numerical stability.

\subsection{Hyperparameters}

The detailed model configurations for different parameter scales are listed in Table~\ref{tab:model-config}. Both model sizes share the same sequence length (2048 tokens) and are trained on 100B tokens, but differ in hidden dimension, gated linear unit (GLU) expansion size, attention head count, and total parameter count.

\begin{table*}[ht]
\centering
\caption{Model configurations for \mname{}}
\label{tab:model-config}
\setlength{\tabcolsep}{5pt}
\renewcommand{\arraystretch}{1.1}
\begin{tabular}{llccccccc}
\toprule
\textbf{Size} & \textbf{Hidden Size} & \textbf{GLU Size} & \textbf{\#Heads} & \textbf{\#Layers} & \textbf{Batch Size} & \textbf{\#Tokens} & \textbf{Seq Length} \\
\midrule
700M & 1536 & 4096 & 16 & 24 & 1M tokens & 100B & 2048 \\
1.3B & 2048 & 5460 & 32 & 24 & 1M tokens & 100B & 2048 \\
\bottomrule
\end{tabular}
\end{table*}
\section{Theoretical Justification}
\label{appendix:justification}
\subsection{Justification for Self-Attention Mechanism.}
\label{justication:attention}
In our complex-valued attention mechanism, we adopt the real part of the Hermitian inner product as the attention score:
\[
S = \operatorname{Re}(\overline{\mathbf{Q}} \mathbf{K}^\top).
\]
This choice is both mathematically principled and practically motivated.

From a geometric perspective, this formulation corresponds to the so-called \emph{Euclidean angle} between complex vectors, as discussed by Scharnhorst~\cite{Scharnhorst:1999angles}. Given a complex vector space $V_\mathbb{C} \cong \mathbb{C}^n$, one can isometrically embed it into a real vector space $V_\mathbb{R} \cong \mathbb{R}^{2n}$ by splitting each complex coordinate into its real and imaginary parts. Under this embedding, the real part of the Hermitian inner product becomes:
\[
\operatorname{Re}(\langle \mathbf{a}, \mathbf{b} \rangle_\mathbb{C}) = \mathbf{a}_\text{re}^\top \mathbf{b}_\text{re} + \mathbf{a}_\text{im}^\top \mathbf{b}_\text{im} = (\mathbf{A}, \mathbf{B})_\mathbb{R},
\]
which is exactly the standard dot product in $\mathbb{R}^{2n}$. Therefore, the real part retains the familiar interpretation of \emph{directional alignment} via projection, analogous to the cosine similarity in real-valued spaces.

More generally, the real part of the Hermitian product appears as the \emph{real component} of the so-called \emph{complex angle} between vectors, defined by:
\[
\cos \Theta_c(\mathbf{a}, \mathbf{b}) = \frac{(\mathbf{a}, \mathbf{b})_\mathbb{C}}{\|\mathbf{a}\| \|\mathbf{b}\|} \in \mathbb{C},
\]
which can be written as $\cos \Theta_c = \rho e^{i\varphi}$, where $\rho = |\cos \Theta_c|$ is the \emph{Hermitian angle} and $\varphi$ is the \emph{pseudo-angle}. Scharnhorst demonstrates that the Euclidean angle serves as a natural projection of this complex-valued structure onto the real line. Specifically, he shows that:
\[
\cos \Theta_c(\mathbf{a}, \mathbf{b}) = \cos \Theta(\mathbf{a}, \mathbf{b}) + i \cos \Theta_K(\mathbf{a}, \mathbf{b}) \sin \Theta(\mathbf{a}, \mathbf{b}),
\]
where $\Theta$ is the Euclidean angle and $\Theta_K$ is the so-called \emph{Kähler angle}. Taking only the real part of the complex angle thus corresponds precisely to using $\cos \Theta(\mathbf{a}, \mathbf{b})$.

This projection not only simplifies implementation by avoiding the need to handle complex-valued scores in softmax, but also preserves the essential geometric information---i.e., how much the vectors are aligned in phase. Since any similarity score in the attention mechanism must ultimately be a real-valued scalar to interface with softmax, $\operatorname{Re}(\overline{\mathbf{Q}} \mathbf{K}^\top)$ emerges as the most natural, interpretable, and structure-preserving choice. It also avoids the ambiguity associated with the phase term $\varphi$, which lacks geometric meaning in projective settings.

In conclusion, using only the real part of the Hermitian inner product offers a geometrically faithful similarity measure, consistent with classical constructions of angles in complex vector spaces~\cite{Scharnhorst:1999angles}, while remaining computationally compatible with attention kernels tailored for modern GPUs.

\subsection{Activation Function in FFN}
\label{justification:act}
The position-wise Feed-Forward Network (FFN) in our architecture adapts the structural principles of modern LLMs like LLaMA to operate entirely within the complex domain. Critically, the choice of the non-linear activation function presents a trade-off between sparsity and gradient smoothness. For instance, the standard Rectified Linear Unit (ReLU) enforces strong sparsity but suffers from a non-differentiable point at the origin. Conversely, functions like SwiGLU offer smoother gradients at the cost of sacrificing this beneficial hard sparsity.To resolve this trade-off, we employ the Squared ReLU (ReLU²) activation function~\cite{zhang2024relu2}, defined for a real input $x$ as:
\[f(x)=\operatorname{ReLU}^2(x) = (\max(0, x))^2.\] 

For a complex pre-activation $\mathbf{Z} = \mathbf{Z}_\text{re} + i\mathbf{Z}_\text{im}$, the function is applied component-wise:
\[f(\mathbf{Z}) = \operatorname{ReLU}^2(\mathbf{Z}_\text{re}) + i \operatorname{ReLU}^2(\mathbf{Z}_\text{im})\]

This function retains the sparsity of ReLU, as its output is zero for the identical set of non-positive inputs. Additionally, the function’s derivative, $\frac{d}{dx}\operatorname{ReLU}^2(x) = 2 \cdot \operatorname{ReLU}(x)$, is continuous across its entire domain, including the origin, which resolves the abrupt gradient change in ReLU and contributes to a more stable optimization landscape. In parallel, the quadratic nature of ReLU² has a valuable inductive bias by enhancing activation contrast. It suppresses weak positive signals while amplifying strong ones, promoting a more decisive and robust feature selection mechanism. Therefore, ReLU² uniquely synthesizes the benefits of high sparsity with smoother gradients and enhanced feature representation, establishing it as an ideal choice for our complex-valued backbone.

\subsection{Derivation of Complex RoPE Embedding}
\label{justification:rope_derivation}

In its original real-valued formulation, RoPE encodes absolute position by applying a rotation matrix to pairs of features in the query and key vectors. Specifically, it pairs adjacent dimensions to simulate 2D rotations. However, in the complex domain, this logic can be implemented more directly and uniformly: a 2D rotation is equivalent to multiplication by a complex exponential of unit modulus, $e^{i\theta}$.

Given a token at position $m$ and hidden dimension index $j$, we define the complex rotary embedding as:
\[
\mathbf{q}'_{m,j} = \mathbf{q}_{m,j} \cdot e^{i m \theta_j}, \quad
\mathbf{k}'_{n,j} = \mathbf{k}_{n,j} \cdot e^{i n \theta_j},
\]
where $\theta_j = \text{base}^{-j/d}$ is a predetermined frequency, and $d$ is the hidden dimension size.

Now consider the Hermitian inner product between the transformed query $\mathbf{q}'_m$ and key $\mathbf{k}'_n$:

\begin{align*}
(\mathbf{q}'_m)^H \mathbf{k}'_n &= (\mathbf{q}_m \odot e^{i m \mathbf{\Theta}})^H (\mathbf{k}_n \odot e^{i n \mathbf{\Theta}}) \\
&= \sum_{j=1}^{d} (\overline{\mathbf{q}_{m,j} e^{i m \theta_j}}) (\mathbf{k}_{n,j} e^{i n \theta_j}) \\
&= \sum_{j=1}^{d} (\overline{\mathbf{q}_{m,j}} \overline{e^{i m \theta_j}}) (\mathbf{k}_{n,j} e^{i n \theta_j}) \\
&= \sum_{j=1}^{d} \overline{\mathbf{q}_{m,j}} \mathbf{k}_{n,j} e^{-i m \theta_j} e^{i n \theta_j} \\
&= \sum_{j=1}^{d} \overline{\mathbf{q}_{m,j}} \mathbf{k}_{n,j} e^{i (n-m) \theta_j}
\end{align*}
where $\odot$ denotes element-wise multiplication and $\mathbf{\Theta}$ is the vector of frequencies $[\theta_1, ..., \theta_d]$. The key properties used are the conjugate of a product ($\overline{ab} = \overline{a}\overline{b}$) and the conjugate of a complex exponential ($\overline{e^{i\phi}} = e^{-i\phi}$).

This result shows that the attention score is modulated by a relative phase shift $e^{i(n-m)\theta_j}$ that depends solely on the position difference $n-m$. Therefore, relative positional information is directly encoded into the inner product in a rotation-equivariant manner.
Unlike the real-valued RoPE approach, which requires pairing dimensions to simulate complex rotation, the complex-valued version applies rotation directly and uniformly to each feature dimension. This leads to a more natural, expressive, and mathematically clean positional encoding mechanism for complex-valued models.

\subsection{Design of \qscheme{}}
\label{justication:qscheme}
Our phase-based quantization scheme is motivated by principles from information theory and complex geometry, offering both representational efficiency and geometric stability.

\subsubsection{Full Information Capacity.}
From an information-theoretic perspective, our goal is to maximize the information encoded within the allocated 2-bit budget. While the ternary set $\{-1, 0, +1\}$ used in BitNet yields approximately 1.58 bits of entropy, our quaternary set $\{\pm1, \pm i\}$ achieves the theoretical maximum of $\log_2(4) = 2$ bits under a uniform distribution. This ensures full utilization of each bit, enhancing the expressiveness of the quantized model.

\subsubsection{Geometric Robustness and Symmetry.}
Our quantization points are chosen as the 4th roots of unity on the complex unit circle. These points are equidistant and symmetrically distributed in the complex plane, maximizing the angular separation between centroids. Such symmetry not only provides geometric robustness and maximal separation in Euclidean distance, but also facilitates uniform treatment of directions, leading to more stable optimization dynamics and improved error resilience.

\subsubsection{Preservation of Directional Information.}
In complex-valued neural networks, the phase primarily encodes directional information, while the magnitude reflects importance. Since we employ per-tensor scaling factors ($\gamma_{re}, \gamma_{im}$) to approximate magnitude, our quantization can focus on accurately preserving phase. This aligns with the intuition that directional information is more critical in many learning tasks, especially in low-bit regimes.

\section{Additional Experimental Results}

\subsection{Quantized Weight Distributions}
\label{app:weight_distribution}

\begin{figure*}[ht]
    \centering
    \begin{subfigure}[t]{0.19\linewidth}
        \centering
        \includegraphics[width=\linewidth]{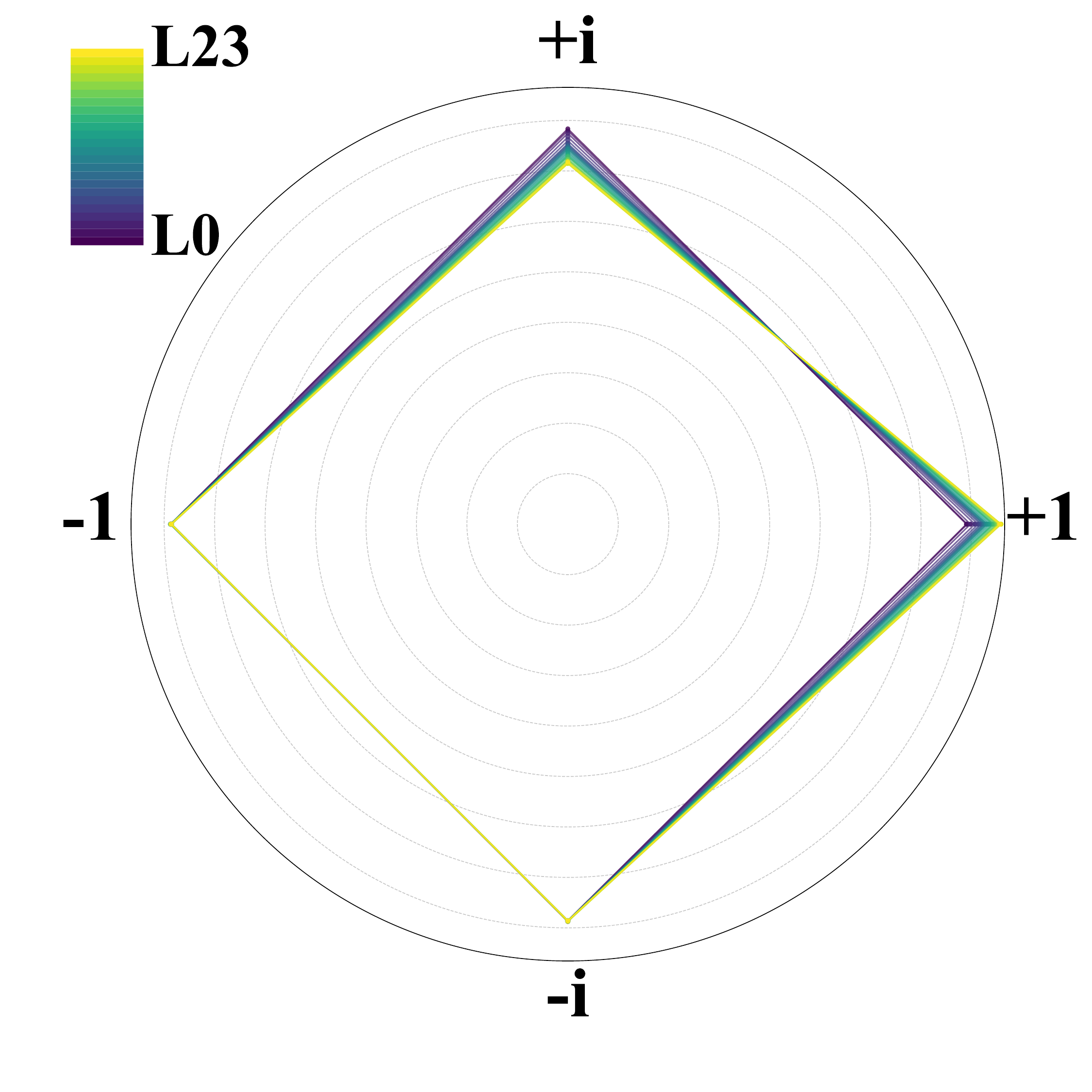}
        \caption{Empirical distribution of quantized weights in $\mathbf{W}_\mathbf{Up}$.}
    \end{subfigure}
    \hfill
    \begin{subfigure}[t]{0.19\linewidth}
        \centering
        \includegraphics[width=\linewidth]{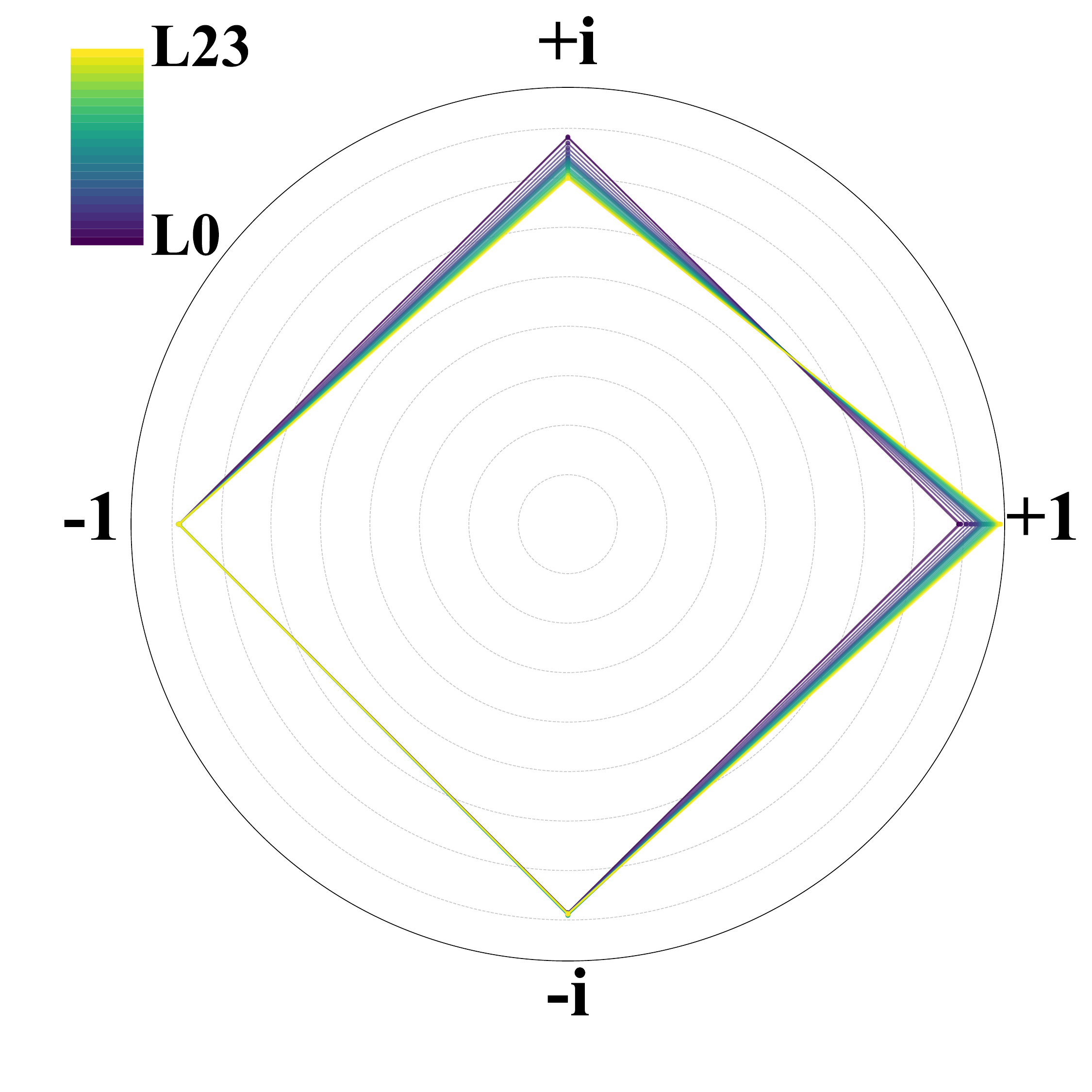}
        \caption{Empirical distribution of quantized weights in $\mathbf{W}_\mathbf{Gate}$.}
    \end{subfigure}
    \hfill\begin{subfigure}[t]{0.19\linewidth}
        \centering
        \includegraphics[width=\linewidth]{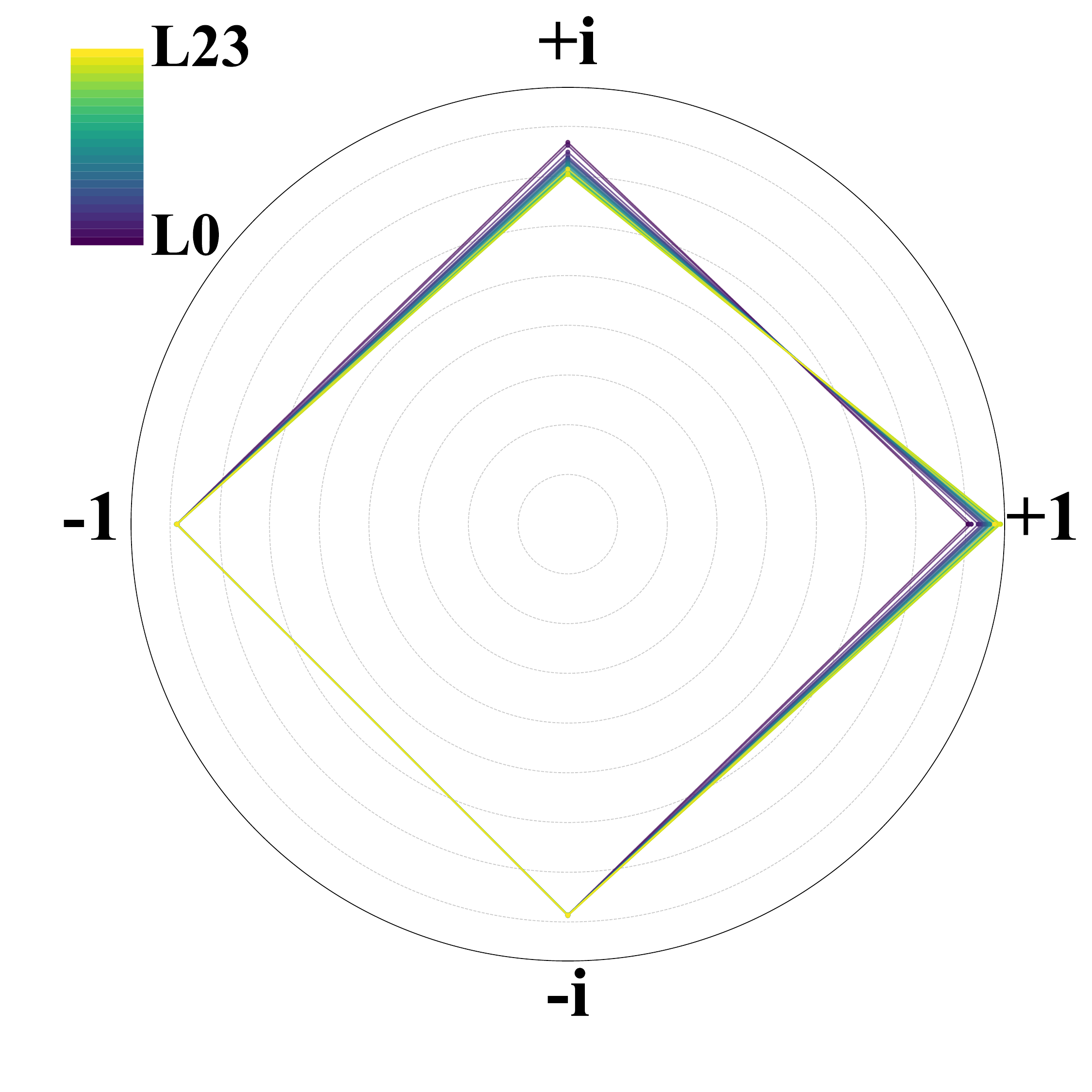}
        \caption{Empirical distribution of quantized weights in $\mathbf{W}_\mathbf{Down}$.}
    \end{subfigure}
    \hfill
    \begin{subfigure}[t]{0.19\linewidth}
        \centering
        \includegraphics[width=\linewidth]{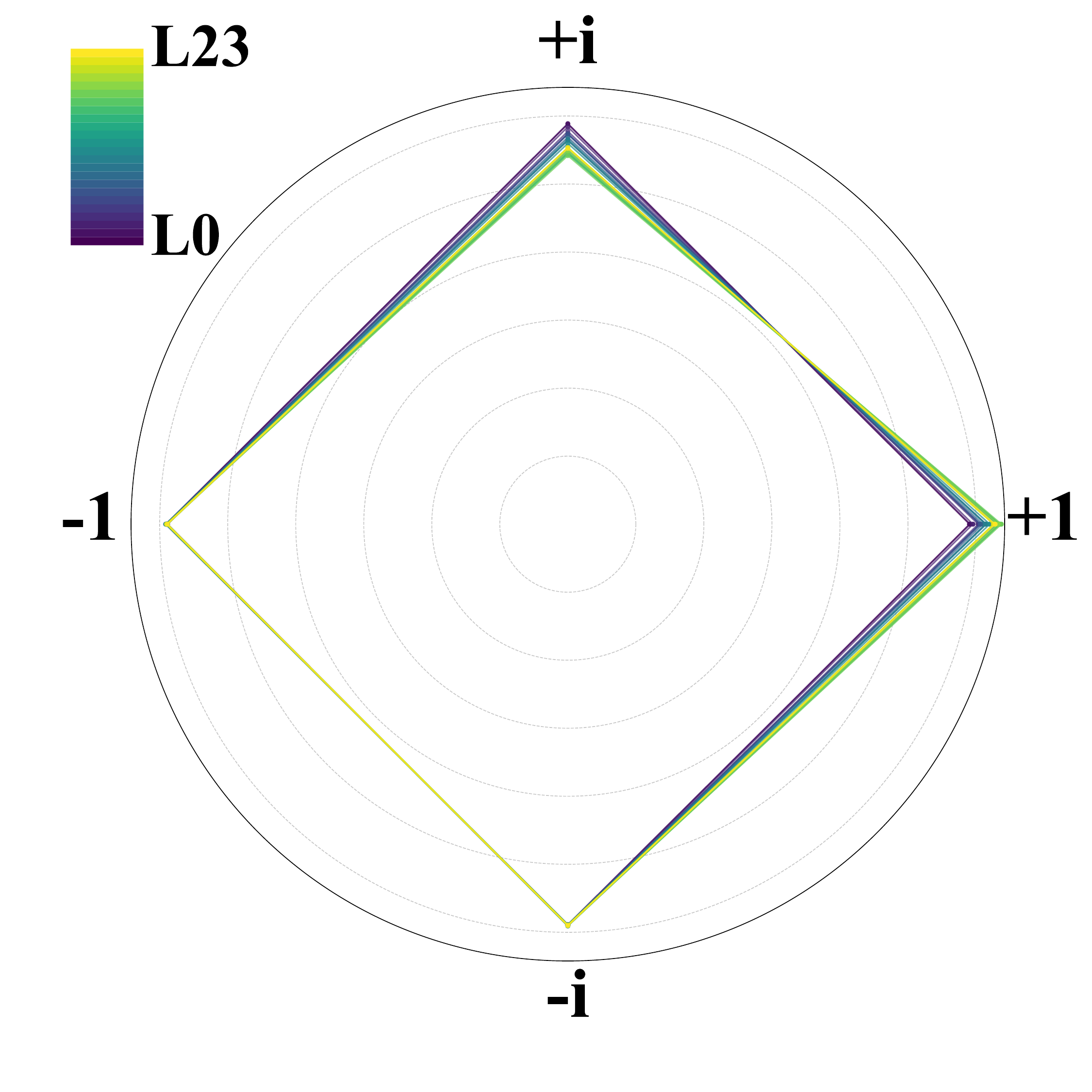}
        \caption{Empirical distribution of quantized weights in $\mathbf{W}_\mathbf{Q}$.}
    \end{subfigure}
    \hfill
    \begin{subfigure}[t]{0.19\linewidth}
        \centering
        \includegraphics[width=\linewidth]{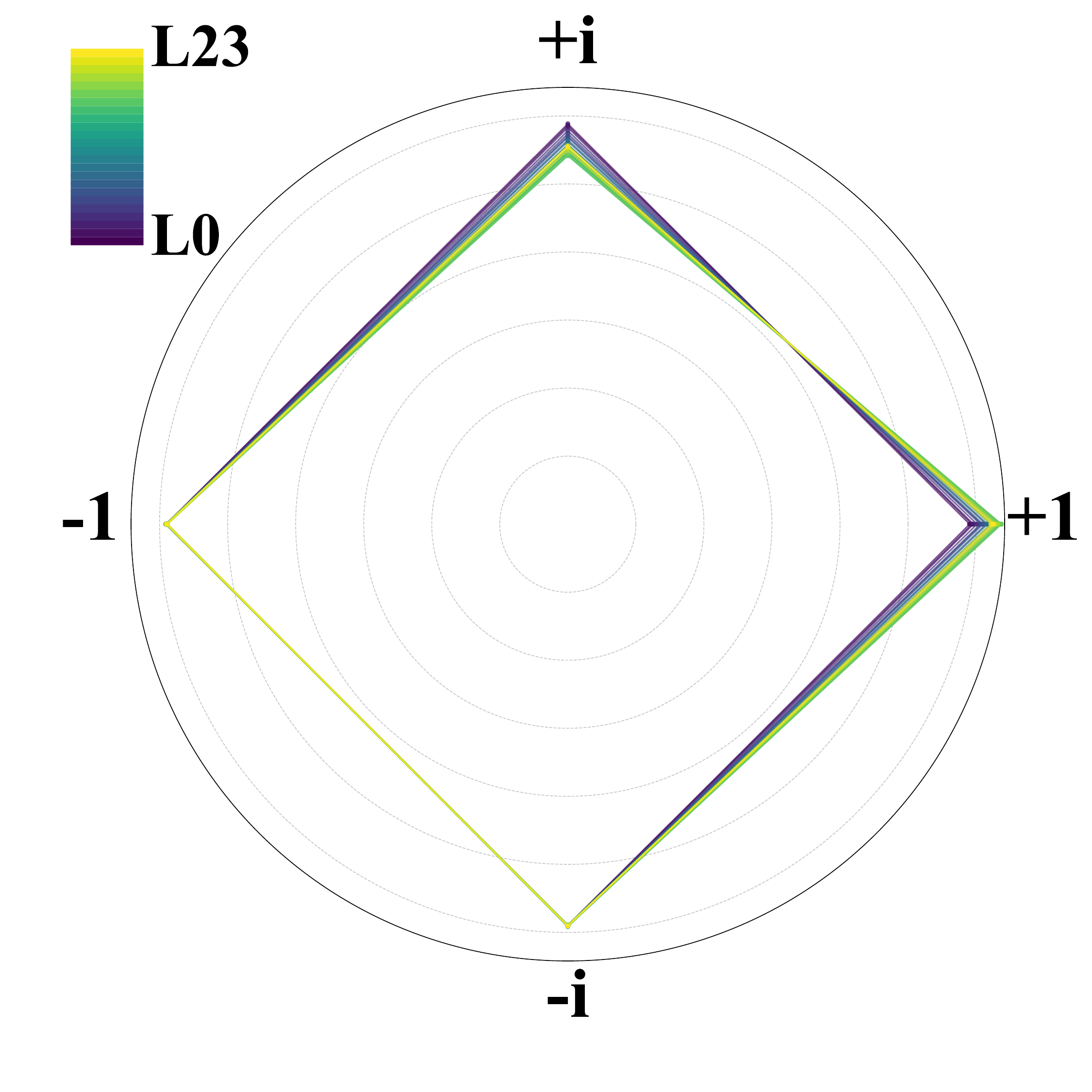}
        \caption{Empirical distribution of quantized weights in $\mathbf{W}_\mathbf{V}$.}
    \end{subfigure}
    \caption{Quantization statistics of weight values in \mname{}.}
    \label{fig:appendix_value_distribution}
\end{figure*}

To complement the main analysis in Section~\ref{subsec_quant_analysis}, we visualize the empirical distributions of quantized weights for all major parameter matrices which is not shown in the main text, including $\mathbf{W}_\mathbf{Q}$ and  $\mathbf{W}_\mathbf{V}$ in the self-attention block, as well as $\mathbf{W}_\text{Up}$, $\mathbf{W}_\text{Gate}$, and $\mathbf{W}_\text{Down}$ in the feed-forward network.
Figure~\ref{fig:appendix_value_distribution} shows the usage frequency of the four complex values $\{\pm1, \pm i\}$ across these modules. All components exhibit balanced or near-uniform distributions, confirming that \mname{} consistently avoids representational collapse and fully exploits the 2-bit codebook throughout the model.

\subsection{Layer-wise Norms}
\label{app:l2norms}

\begin{figure*}[ht]
    \centering
    \begin{subfigure}[t]{0.24\linewidth}
        \centering
        \includegraphics[width=\linewidth]{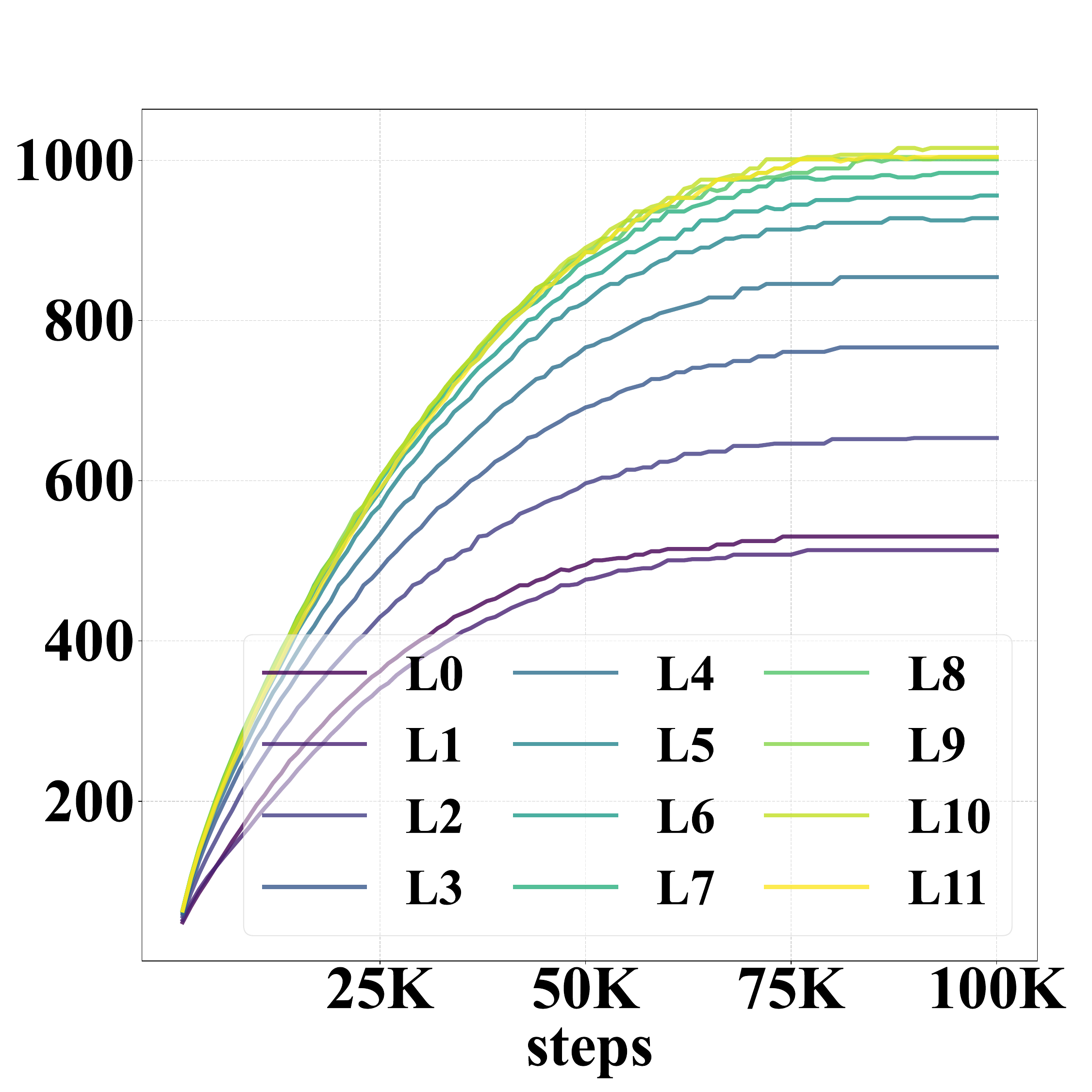}
        \caption{Up projection $\ell_2$ norm across layer 0 to layer 11.}
    \end{subfigure}
    \hfill
    \begin{subfigure}[t]{0.24\linewidth}
        \centering
        \includegraphics[width=\linewidth]{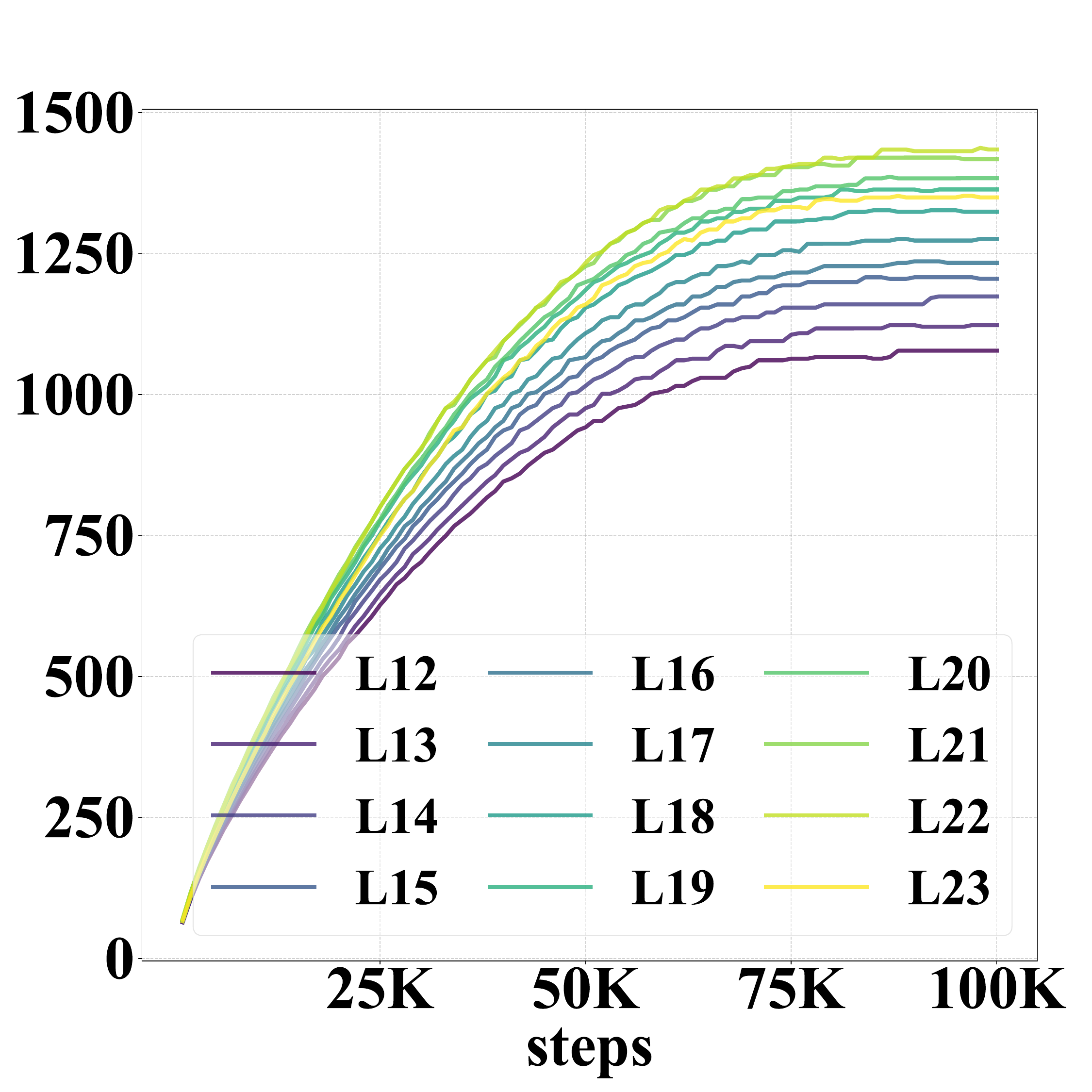}
        \caption{Up projection $\ell_2$ norm across layer 12 to layer 23.}
    \end{subfigure}
    \hfill
    \begin{subfigure}[t]{0.24\linewidth}
        \centering
        \includegraphics[width=\linewidth]{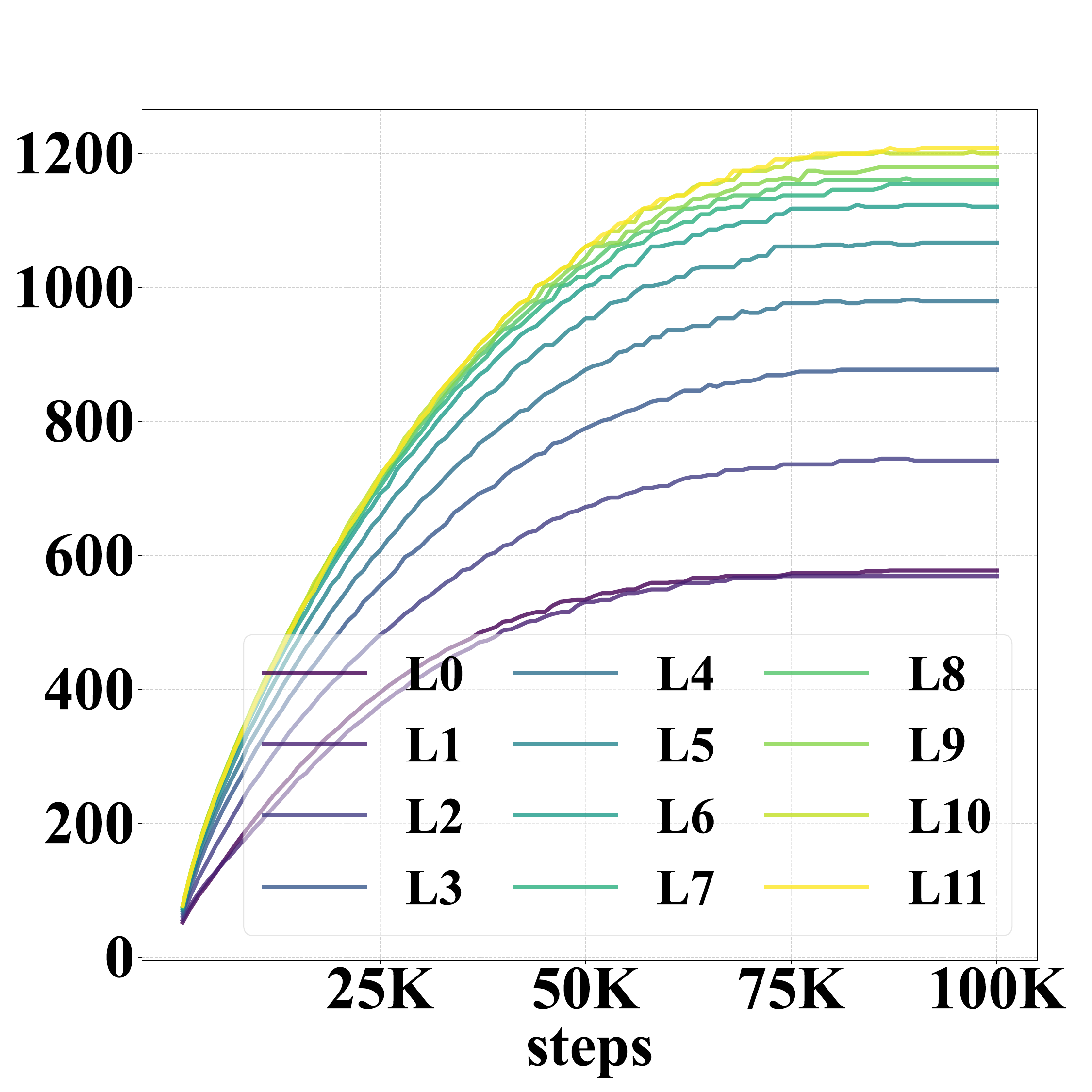}
        \caption{Gate projection $\ell_2$ norm across layer 0 to layer 11.}
    \end{subfigure}
    \hfill
    \begin{subfigure}[t]{0.24\linewidth}
        \centering
        \includegraphics[width=\linewidth]{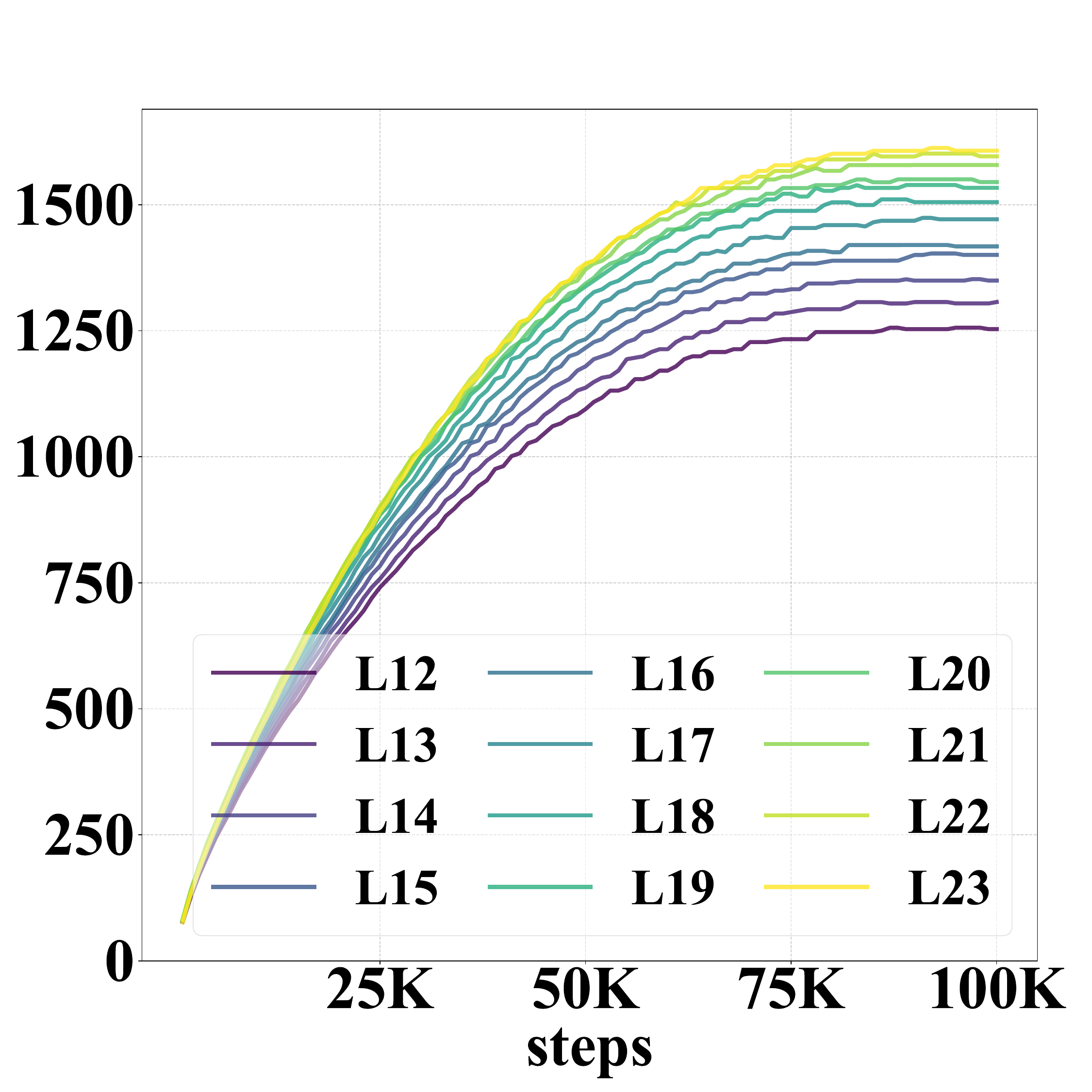}
        \caption{Gate projection $\ell_2$ norm across layer 12 to layer 23.}
    \end{subfigure}
    \vspace{0.5em}

    \begin{subfigure}[t]{0.24\linewidth}
        \centering
        \includegraphics[width=\linewidth]{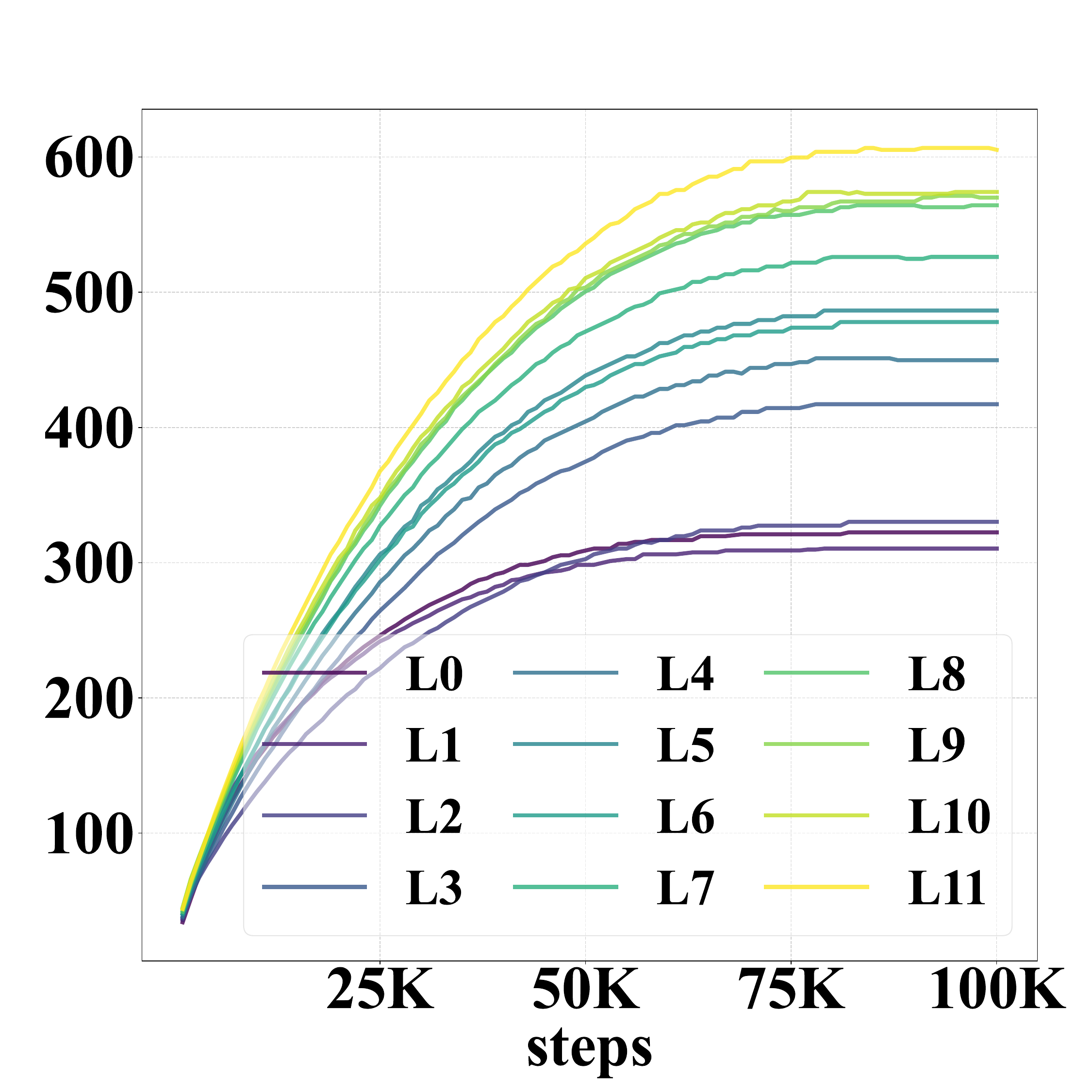}
        \caption{Q projection $\ell_2$ norm across layer 0 to layer 11.}
    \end{subfigure}
    \hfill
    \begin{subfigure}[t]{0.24\linewidth}
        \centering
        \includegraphics[width=\linewidth]{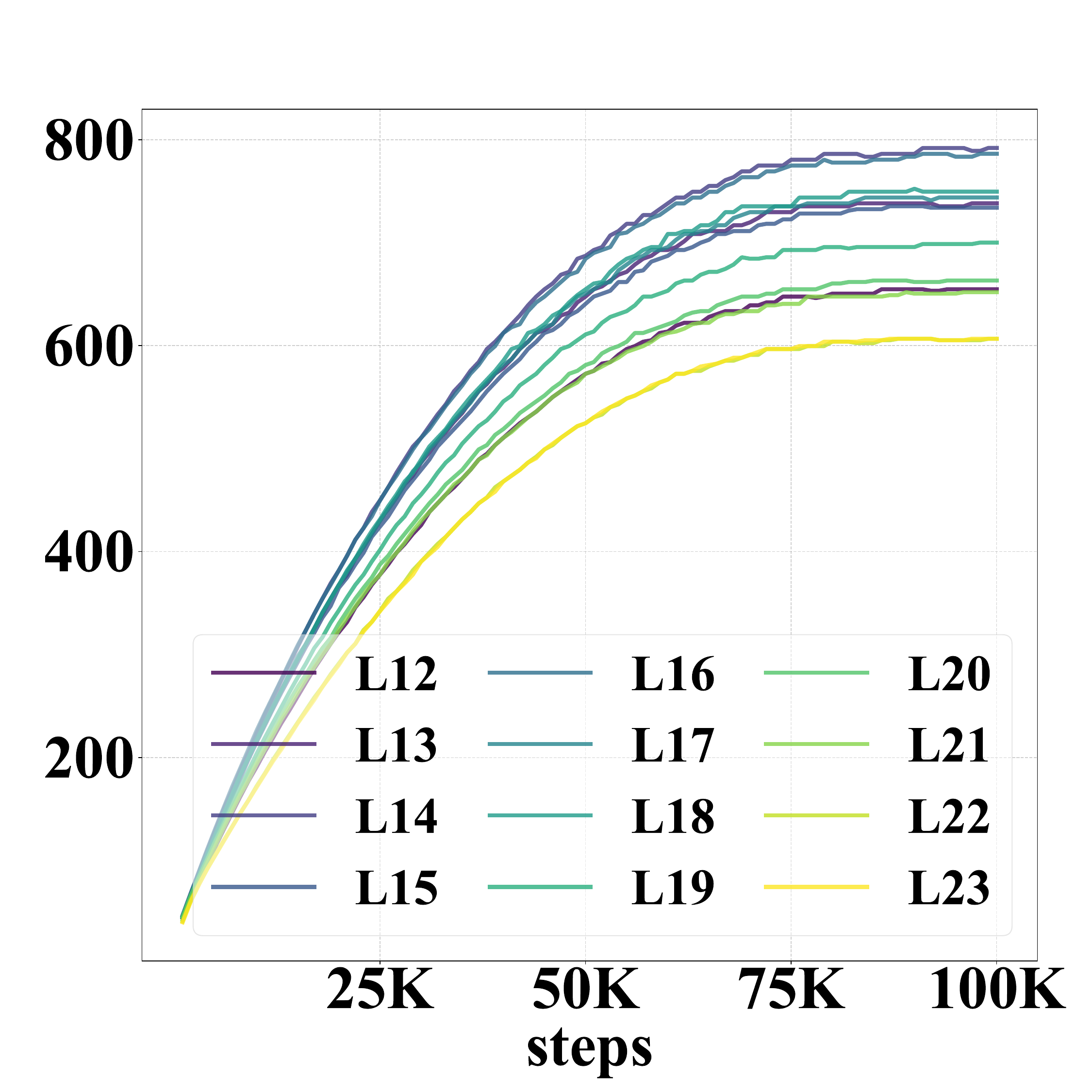}
        \caption{Q projection $\ell_2$ norm across layer 12 to layer 23.}
    \end{subfigure}
    \hfill
    \begin{subfigure}[t]{0.24\linewidth}
        \centering
        \includegraphics[width=\linewidth]{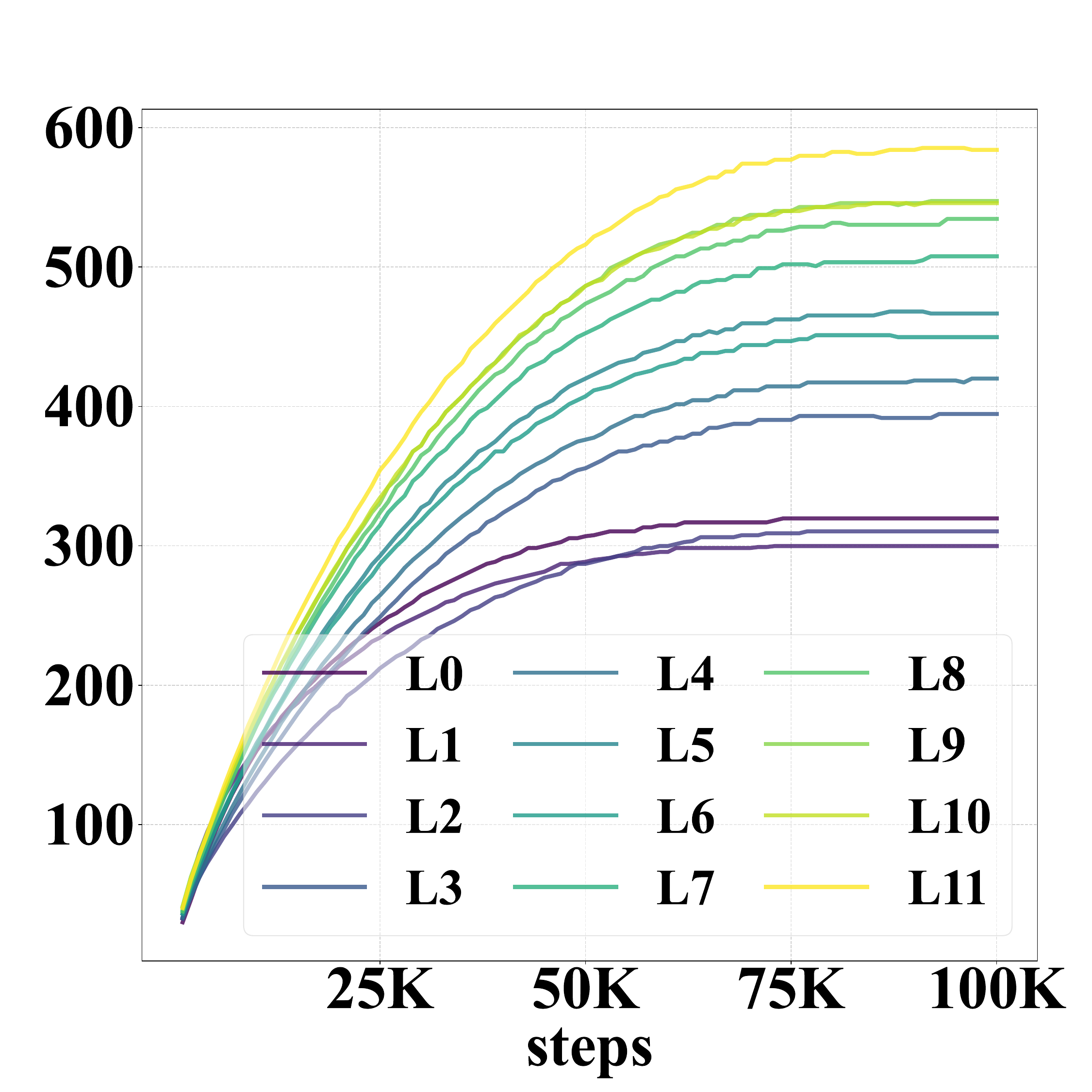}
        \caption{V projection $\ell_2$ norm across layer 0 to layer 11.}
    \end{subfigure}
    \hfill
    \begin{subfigure}[t]{0.24\linewidth}
        \centering
        \includegraphics[width=\linewidth]{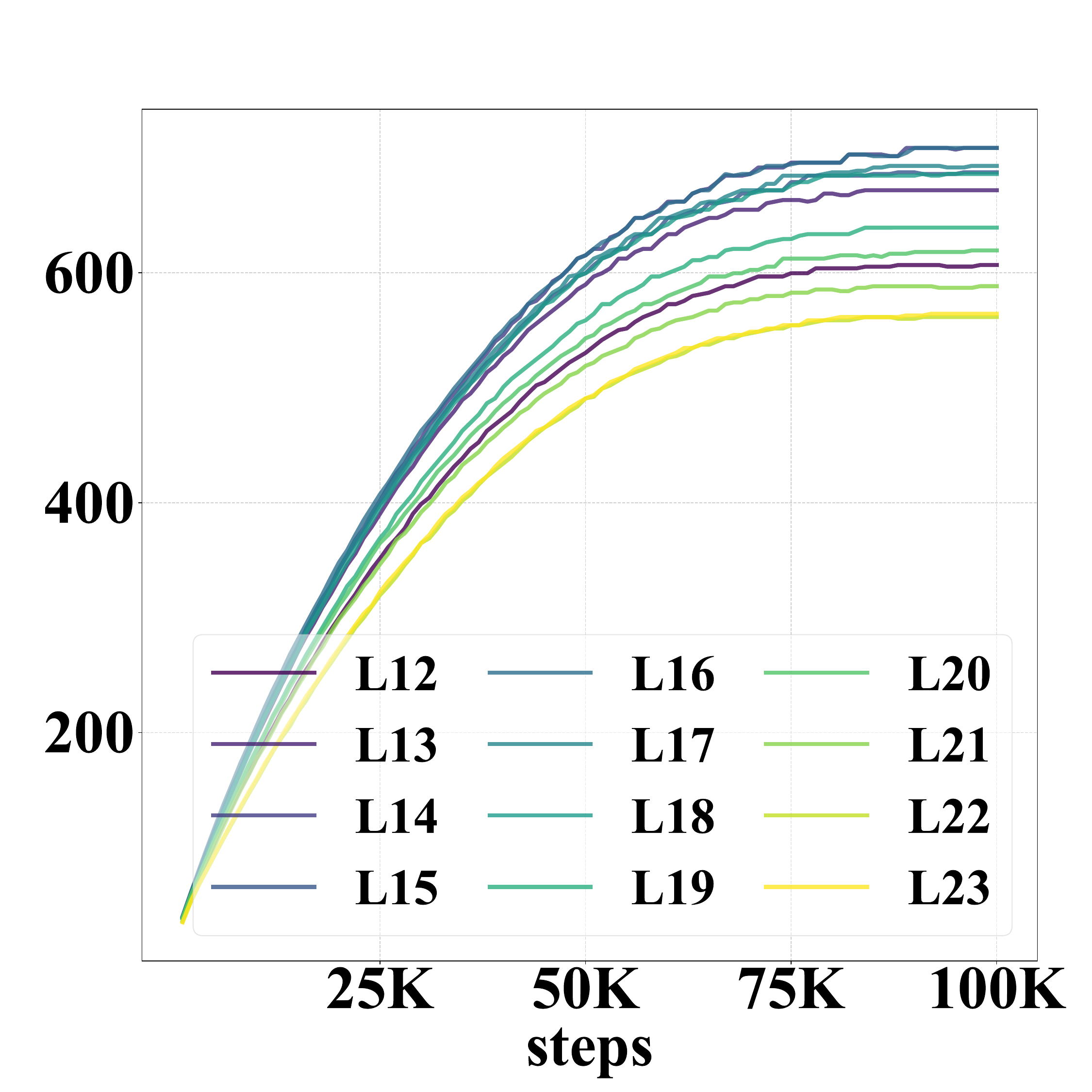}
        \caption{K projection $\ell_2$ norm across layer 12 to layer 23.}
    \end{subfigure}
    \vspace{0.5em}

    \begin{subfigure}[t]{0.24\linewidth}
        \centering
        \includegraphics[width=\linewidth]{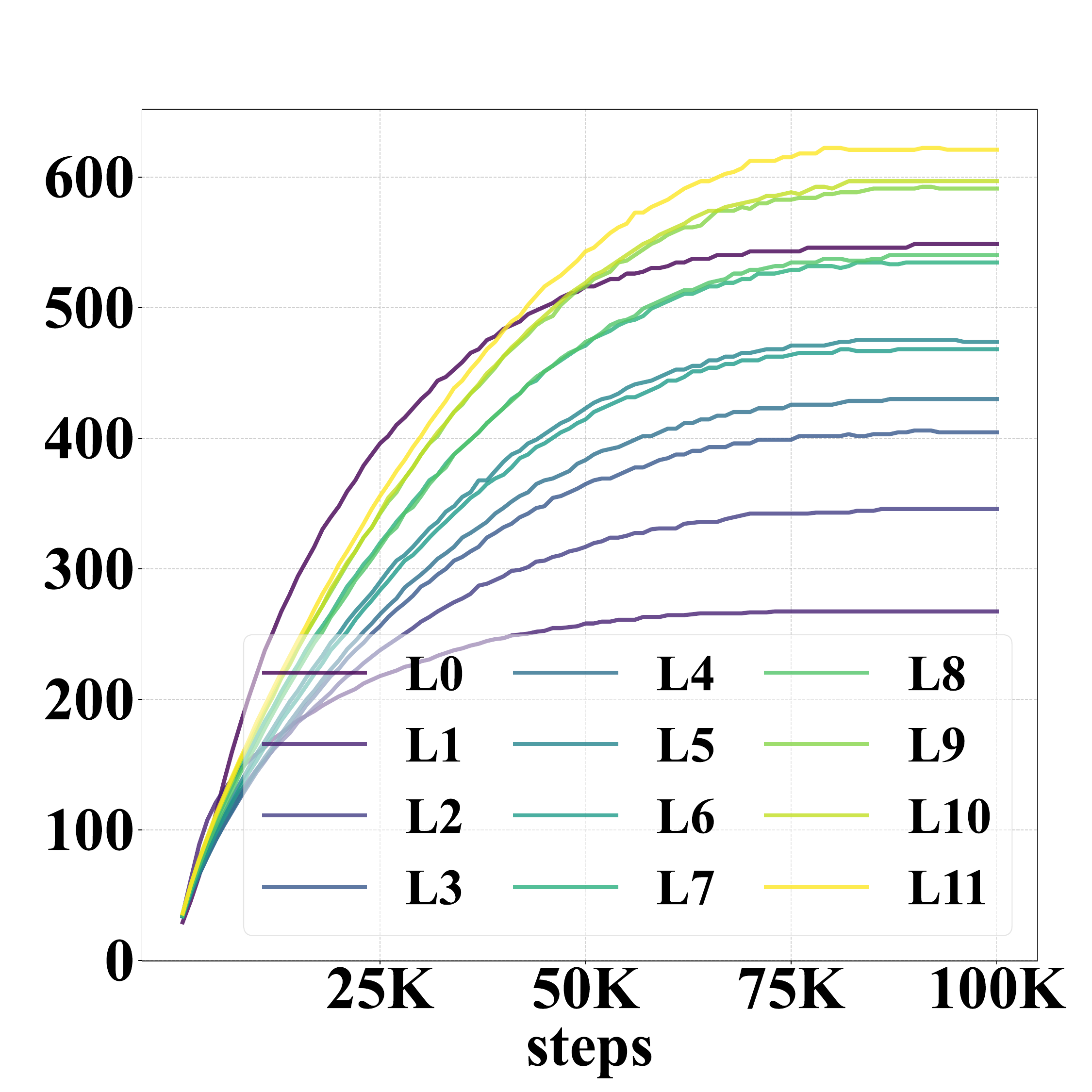}
        \caption{V projection $\ell_2$ norm across layer 0 to layer 11.}
    \end{subfigure}
    \hfill
    \begin{subfigure}[t]{0.24\linewidth}
        \centering
        \includegraphics[width=\linewidth]{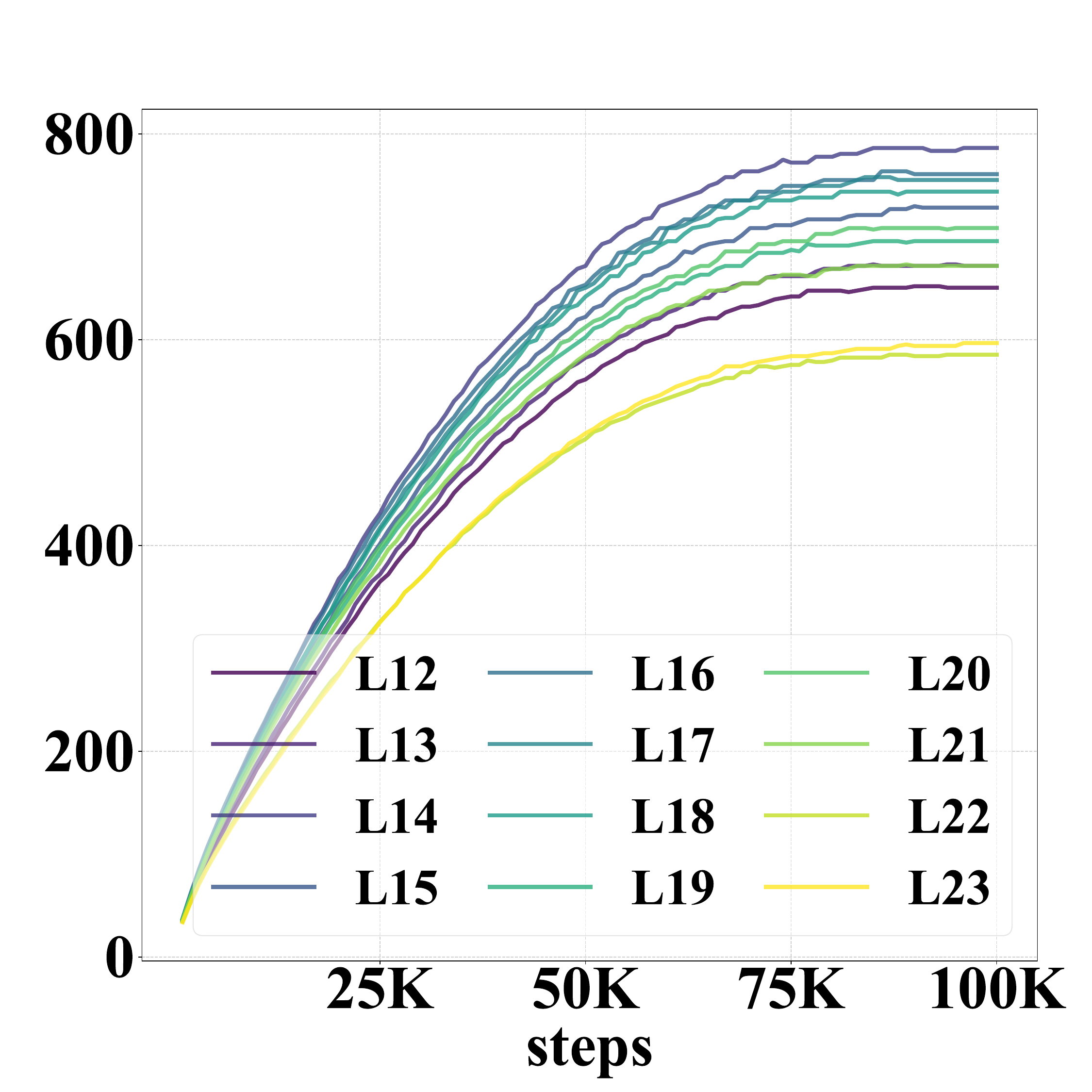}
        \caption{V projection $\ell_2$ norm across layer 12 to layer 23.}
    \end{subfigure}
    \hfill
    \begin{subfigure}[t]{0.24\linewidth}
        \centering
        \includegraphics[width=\linewidth]{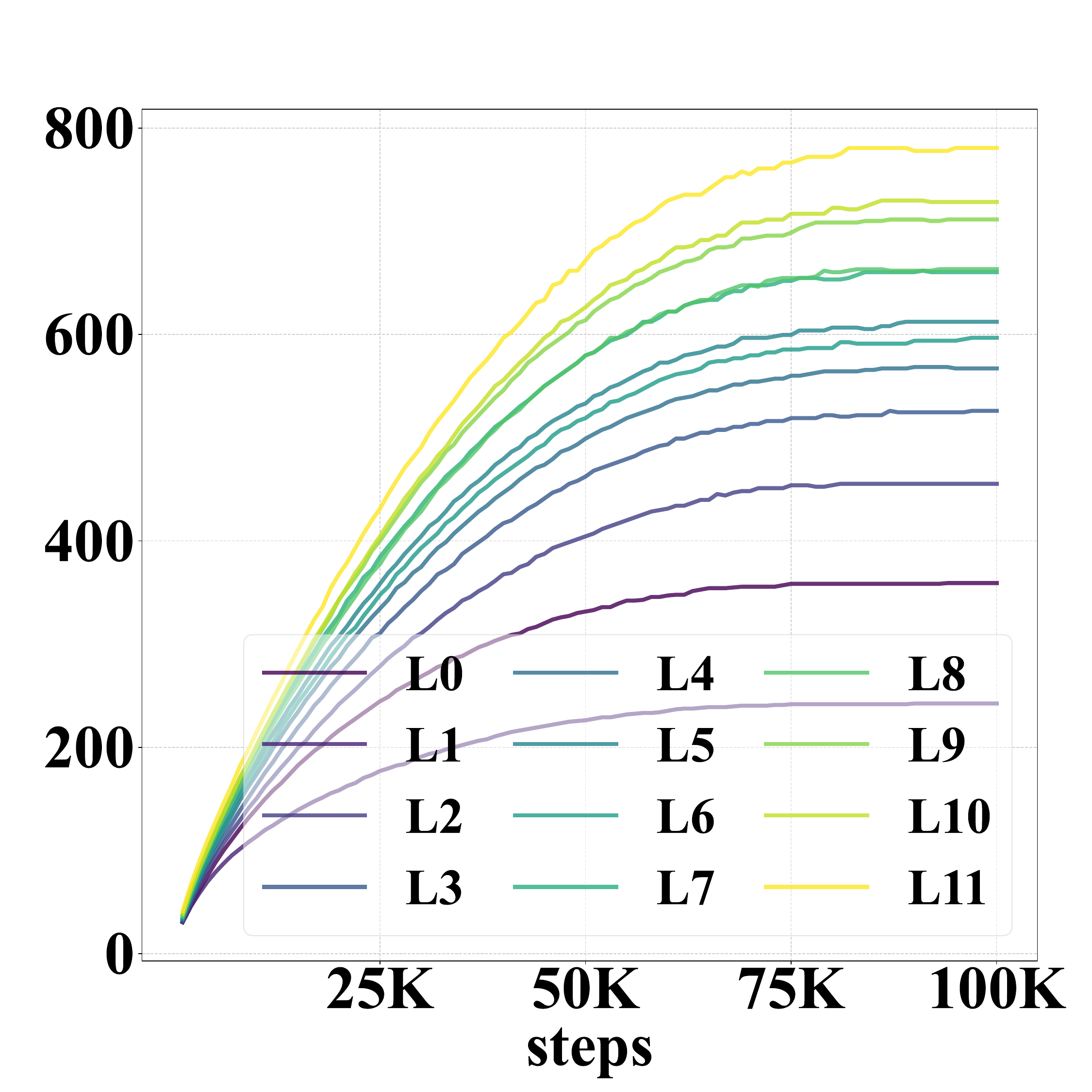}
        \caption{O projection $\ell_2$ norm across layer 0 to layer 11.}
    \end{subfigure}
    \hfill
    \begin{subfigure}[t]{0.24\linewidth}
        \centering
        \includegraphics[width=\linewidth]{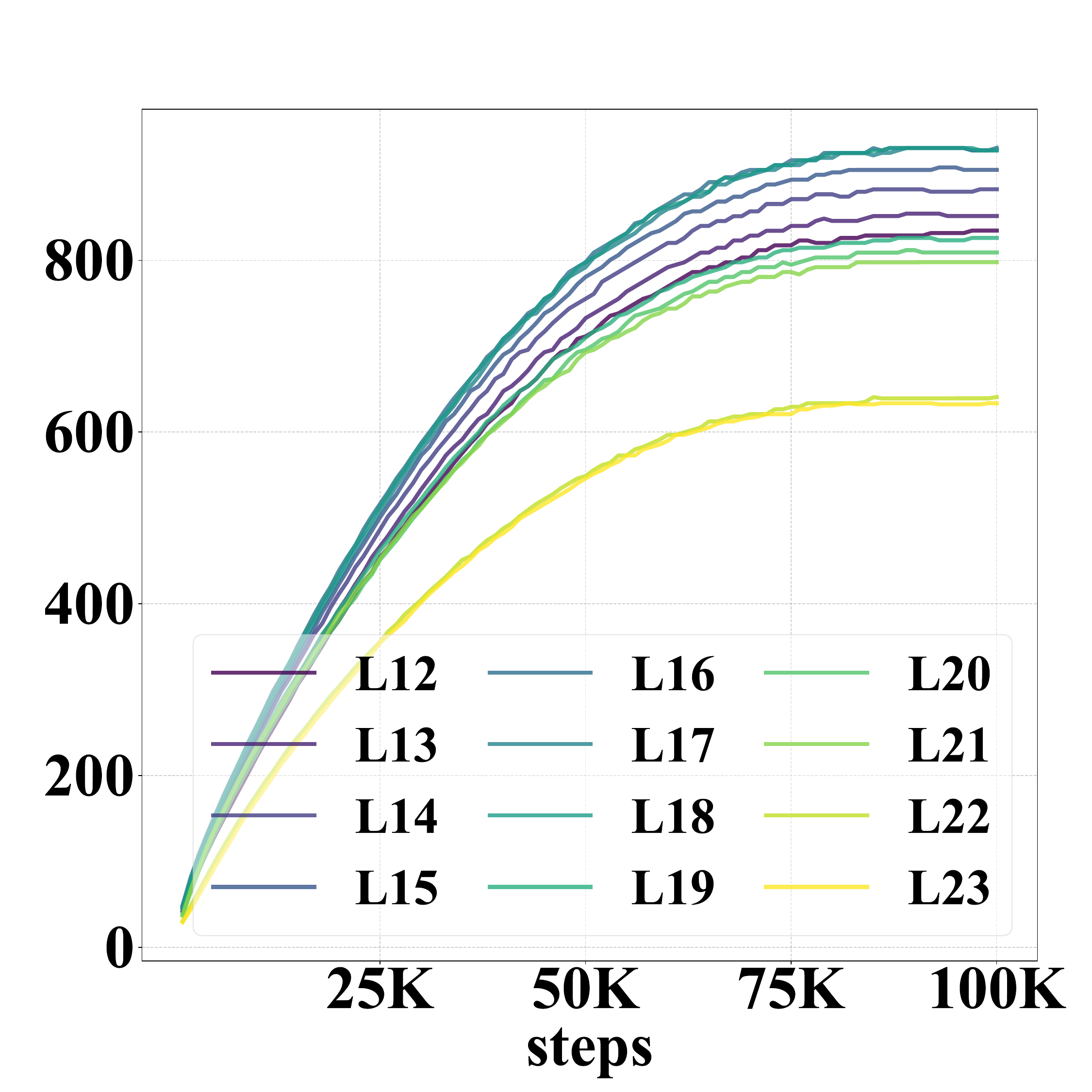}
        \caption{O projection $\ell_2$ norm across layer 12 to layer 23.}
    \end{subfigure}
    
    \caption{Layer-wise $\ell_2$ norm of complex-valued quantized weights in \mname{}.}
    \label{fig:appendix_l2_norms}
\end{figure*}

To complement the analysis in Section~\ref{subsec_quant_analysis}, we provide the $\ell_2$ norms of all quantized weight matrices across layers. As shown in Figure~\ref{fig:appendix_l2_norms}, these components exhibit similarly stable norm distributions, further confirming that our quantization scheme consistently preserves the scale structure across the entire model.

